\documentclass{article}

\usepackage[main, final]{neurips_2026}

\usepackage[utf8]{inputenc} % allow utf-8 input
\usepackage[T1]{fontenc}    % use 8-bit T1 fonts
\usepackage{xcolor}         % colors
\definecolor{Highlight}{rgb}{0.12,0.49,0.85}
\usepackage[breaklinks,colorlinks,linkcolor=purple,citecolor=Highlight]{hyperref}
\usepackage{url}            % simple URL typesetting
\usepackage{booktabs}       % professional-quality tables
\usepackage{amsfonts}       % blackboard math symbols
\usepackage{nicefrac}       % compact symbols for 1/2, etc.
\usepackage{amsmath}
\usepackage{pifont}
\usepackage{enumitem}
\usepackage{multirow}
\usepackage{longtable}
\usepackage{graphicx}
\usepackage{needspace}

\usepackage{xspace} % 使用xspace可以自动智能地判断后面是否需要空格
\newcommand{\NAME}{{\fontfamily{lmtt}\selectfont \textbf{Timeflies}}\xspace}
\newcommand{\BenchName}{{\fontfamily{lmtt}\selectfont \textbf{Shadow}}\xspace}

\usepackage[capitalize]{cleveref}
\crefname{section}{Sec.}{Secs.}
\Crefname{section}{Section}{Sections}
\Crefname{table}{Table}{Tables}
\crefname{table}{Tab.}{Tabs.}

\newcommand{\appref}[1]{App.~\ref{#1}}

\usepackage{threeparttable}
\usepackage{multirow, multicol}
\usepackage{graphicx}
\usepackage{caption}
\usepackage{subfigure}

\usepackage{amssymb}% http://ctan.org/pkg/amssymb
\usepackage{pifont}% http://ctan.org/pkg/pifont
\newcommand{\cmark}{\ding{51}}%

\newcommand{\icon}{\raisebox{-2pt}{\includegraphics[width=1.2em]{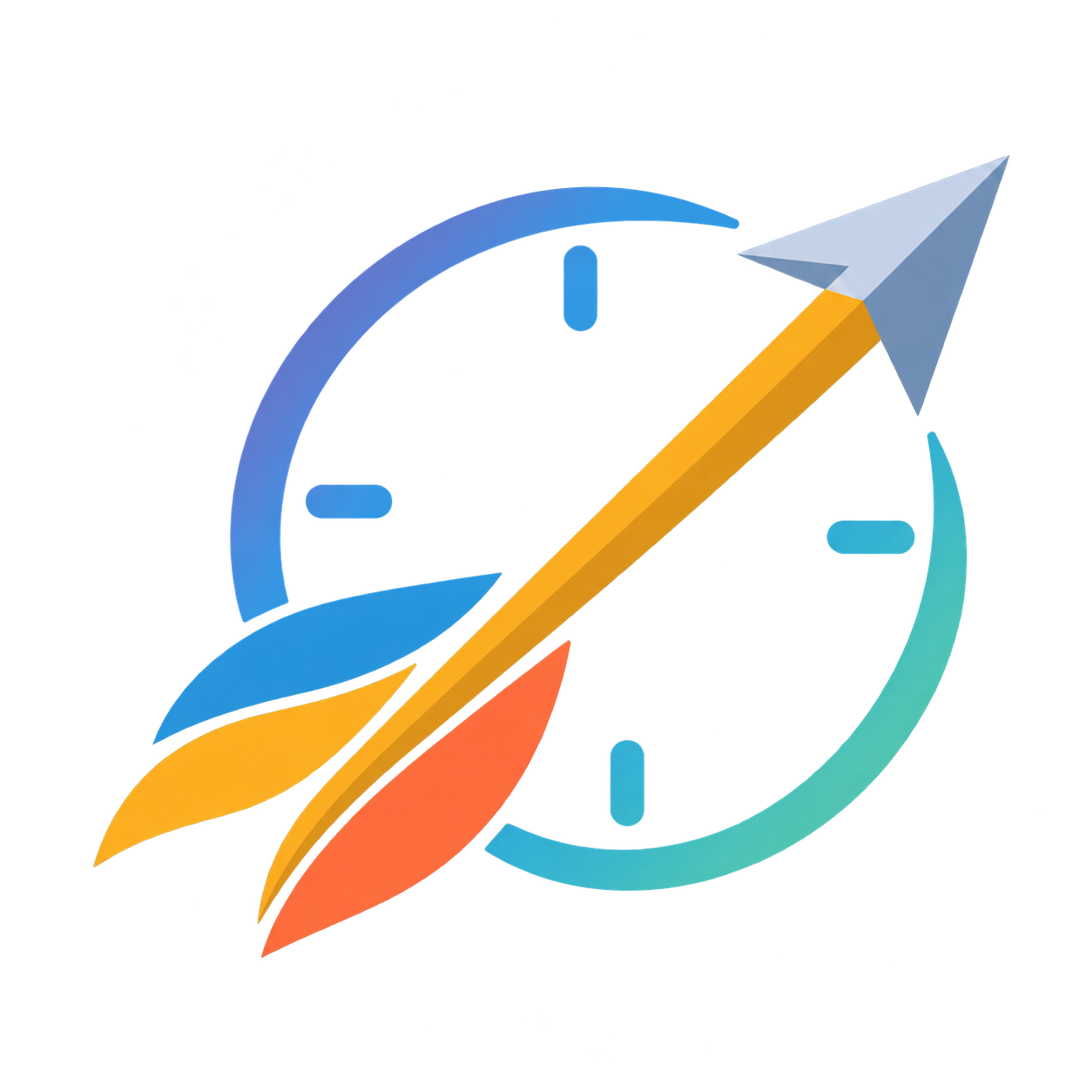}}\xspace}

\title{\icon\ Existence Precedes Value: Joint Modeling of Observational Existence and Evolving States in\\Time Series Forecasting
}

\author{%
  \textbf{Yifan Hu$^{1,2,}$\thanks{Equal contribution.\qquad$^\dagger$ Corresponding author.}\ ,\ \ Hongzhou Chen$^{1,*}$,\ \ Peiyuan Liu$^{2,*}$,\ \ Yiding Liu$^{1}$,}\\
  \textbf{Zewei Dong$^{1,\dagger}$,\ \ Jiang-Ming Yang$^{1}$}\\
  $^1$ Ant International \qquad $^2$ Tsinghua University \\
  \texttt{\{hyf476357, hongzhou.chz, zewei.dong\}@ant-intl.com} \\
}

\begin{document}

\maketitle

\begin{abstract}
Real-world time series are often highly incomplete and irregular due to sensor dormancy, transmission delays, and event-driven sampling, making reliable forecasting fundamentally challenging. Existing methods have evolved from impute-then-forecast pipelines to continuous-time models such as Neural ODEs and continuous-time graph networks. While these approaches improve the modeling of historical irregularity, they still rely on an implicit oracle assumption at inference time: the timestamps of future valid observations are presumed to be known in advance. This assumption limits practical relevance, since in many real systems the more fundamental question is not only what the future value will be, but also whether a valid observation will occur at all. In this paper, we propose \NAME, a unified framework that reformulates forecasting as a joint problem of future observability inference and value estimation. To explicitly model the interaction between observation dynamics and state evolution, \NAME adopts an observation stream and a value stream, coupled through three dedicated modules for reliability-aware embedding, observation-guided dependency modeling, and joint prediction. We further construct \BenchName, a benchmark that combines natural missingness from public datasets with real-world industrial data, and introduce the Observation-Value Joint Entropy (OVJE) metric to comprehensively evaluate this coupled predictability. Extensive experiments show strong joint forecasting performance and competitive value accuracy. The additional value-forecasting benefit depends on the observation mechanism and is not guaranteed under random or value-independent missingness. Code and dataset are available in \url{https://github.com/ant-intl/Timeflies}.

  % \NAME, \textbf{Time} series \textbf{f}orecasting via \textbf{l}earning \textbf{i}nteractive \textbf{e}xistence and \textbf{s}tates

  % \BenchName, \textbf{G}enuine \textbf{L}arge-scale \textbf{I}rregular \textbf{M}issing \textbf{P}attern \textbf{S}equential \textbf{E}valuation 
\end{abstract}

\section{Introduction}

\begin{quote}
    ``\textit{Time flies over us, but leaves its shadow behind.}''
\text{--- Nathaniel Hawthorne}
\end{quote}

\begin{figure}[!t]
  \begin{center}
    \includegraphics[width=\textwidth]{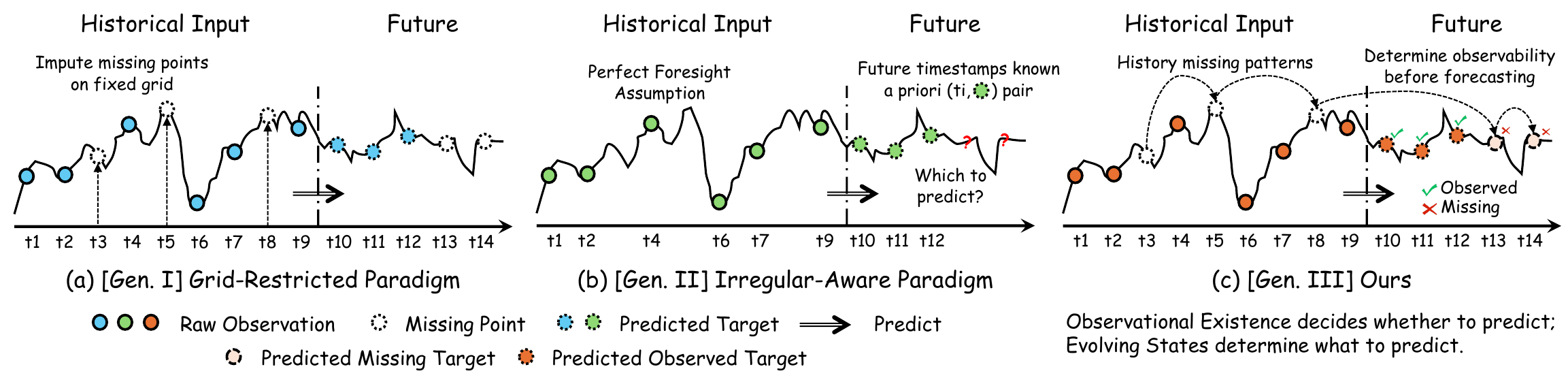}
  \end{center}
  \vspace{-5pt}
  \caption{Evolution of Forecasting Paradigms for Time Series. 
  (a) \textbf{Grid-Restricted Paradigm}: Treats missingness as a defect and employs a rigid impute-then-forecast approach for aligned time slots.
  (b) \textbf{Irregular-Aware Paradigm}: Models continuous latent trajectories from irregular historical observations. However, it requires prior knowledge of future valid timestamps during inference.
  (c) \textbf{Observation-State Joint Paradigm} (Ours): Elevates forecasting to a cascaded dual-track task. It explicitly decodes historical observational regularity to jointly predict future observability and value evolution. This formulation provides complementary information for downstream decisions.}
  \vspace{-5pt}
  % \caption{Evolution of forecasting paradigms for time series.
  % (a) \textbf{Grid-Restricted Paradigm}: Treats missingness as a defect and follows an impute-then-forecast pipeline on aligned grids.
  % (b) \textbf{Irregular-Aware Paradigm}: Models continuous latent dynamics from irregular histories, but assumes future valid timestamps are known as priors at inference.
  % (c) \textbf{Observation-State Joint Paradigm} (Ours): Predicts future observability and values jointly by decoding historical observation regularity, ensuring practical relevance in real-world deployment.}
  % \vspace{-20pt}
\label{fig:intro}
\end{figure}

In real-world time series systems, the generation of observations is often characterized by significant incompleteness and non-uniformity, due to factors such as sensor dormancy, transmission latencies, and event-driven selective sampling~\cite{du2023saits, yang2025revisiting, nie2024imputeformer}. Despite these inherent irregularities, achieving precise forecasting remains mission-critical in these areas. 
To this end, time series forecasting (TSF) has evolved through two major paradigms as shown in \cref{fig:intro}. 
The first generation treated missing data as a defect to be eliminated, using ``impute-then-forecast'' strategies that forced data onto a rigid, uniform grid~\cite{informer,pdf}. 
To overcome these limitations, a second generation of models has emerged, including Neural Ordinary Differential Equations (Neural ODEs) and continuous-time graph networks. These frameworks treat time as a continuous variable, enabling state estimation at arbitrary timestamps and significantly improving the modeling of non-uniform historical sequences~\cite{zhangirregular2024,luo2025hipatch}.

However, a critical yet overlooked limitation remains: while second-generation models excel at fitting historical irregularities, they operate under the assumption of \textit{\textbf{perfect foresight}} during inference. 
Specifically, these models assume that the timestamps of future valid observations are known in advance~\cite{hornek2025value,liu2026rethinking}. In real-world systems, however, it is often unclear whether a valid observation will even occur at a specific future moment, as this depends on the underlying logic of the system~\cite{weather_forecast,hu2025fintsb,traffic_flow}. Predicting a numerical value alone does not indicate whether a future measurement will be available to verify it; jointly predicting observability can therefore inform sensing and decision-making.

Missingness can provide a temporal signal reflecting the system's behavioral patterns. 
Fundamentally, real-world time series are driven by two coupled processes: \ding{182} \textit{\textbf{Observational Existence}}, which determine whether that state is captured as valid data, and \ding{183} \textit{\textbf{Evolving States}}, which dictate the evolution of physical or logical attributes.
\textit{Over time, the value of observational existence has been underestimated}~\cite{liu2025heterogeneous,yu2025ginar+,yang2026from, timegs}.
However, the physical manifestation of these dynamics, or the trajectory of historical missingness, is rarely random.
Rather, it encodes structural priors such as periodic sensor cycles, load-driven sparsity, or state-triggered sampling. Consequently, historical missing patterns provide vital context for predicting the future observability of the system~\cite{zamanian2024analysis, liu2026falconx, xia2026falcon2}.

We therefore formulate forecasting as joint prediction of values and future observability. These complementary outputs address two questions:
\begin{itemize}[leftmargin=*]
    \item[\ding{172}] \textit{\textbf{Future Observational Existence.}} Based on historical missing trajectories, the model transforms missingness from a noise factor into a structured prior for predicting future observations.
    \item[\ding{173}] \textit{\textbf{Observation-Aware Evolving States.}} The model predicts future values using the available history, while the observation forecast characterizes whether those values will be measured.
\end{itemize}
By explicitly modeling the dependency between historical missingness and future observational states, the captured observational regularity can provide additional context for numerical regression when it carries state-relevant information~\cite{yuan2022data, hu2026tomorrow}.

Based on the above motivation, we propose \textbf{\NAME}, a unified forecasting framework that reformulates time series prediction as a joint process of existence inference and value estimation. 
To operationalize this dual-dynamic paradigm, our architecture explicitly captures these coupled processes through a symmetric design: a dedicated \textit{value stream} and an \textit{observation stream}.
% To capture the true data generation process, we meticulously uncouple the temporal dynamics into two symmetric pathways: the underlying \textit{evolving states} and their \textit{observational existence}. 
% Correspondingly, our \NAME explicitly models these dynamics through two dedicated streams: a \textit{value stream} and an \textit{observation stream}.
Specifically, the architecture is driven by three synergistic modules.
\textit{Reliability-Gated Patch Embedding} calculates a deterministic reliability score based on the observation ratio and missing intervals. It then explicitly attenuates sparse, noise-dominated patches, ensuring robust semantic-rich representations enter the subsequent network.
\textit{Cross-track Conditioned Transformer} introduces an observation-conditioned attention mechanism that explicitly incorporates learned historical missing patterns from the observation stream as a structural prior to guide the value stream.
\textit{Bifurcated Observation-Evolution Head} simultaneously determines future observability probabilities and regresses expected state values under a joint optimized objective, aligning with our ``existence precedes value'' philosophy.

Furthermore, to rigorously validate the capability of our framework in realistic deployments, we curate \textbf{\BenchName}, a benchmark combining natural missing gaps from open-source datasets with real-world industrial data.
Crucially, moving beyond traditional marginal error metrics, we argue that a holistic evaluation must be grounded in a joint probability perspective. Therefore, we introduce the Observation-Value Joint Entropy (\textbf\textit{{OVJE}}) metric, designed to explicitly quantify the synergy between accurate existence inference and precise state estimation. 
Evaluated under this unified criterion on the \BenchName benchmark, \NAME achieves strong aggregate joint performance, supporting future observability as an explicit forecasting target.

In a nutshell, our contributions are as follows:
\begin{itemize}[leftmargin=*,itemsep=-0.1em]
    \item \textit{\textbf{Joint Forecasting Formulation.}} We treat future observational existence as an explicit forecasting target alongside value prediction, relaxing the assumption that future valid timestamps are supplied. The observation forecast provides additional information; improvements in value accuracy depend on the observation mechanism.
    \item \textit{\textbf{Technical advancement.}} We propose \NAME, a framework that explicitly decomposes the time series into two interactive streams: \textit{observation stream} and \textit{value stream}. By extracting historical observation rhythms as dynamic context, the architecture enables the existence patterns to explicitly guide and refine the state predictions. The contribution is their task-specific coupling within a joint prediction formulation; attention, focal loss, and residual fusion are not individually new components.
    \item \textit{\textbf{Empirical and Practical Utility.}} To bridge the evaluation gap in realistic scenarios, we construct \BenchName, a rigorous benchmark combining open-source and real-world industrial datasets with natural missing patterns. Moreover, we propose the Observation-Value Joint Entropy (OVJE) metric to holistically evaluate the coupled capability of existence inference and value estimation. Experiments support strong aggregate joint performance, with data-dependent gains in value forecasting.
\end{itemize}

\vspace{-4pt}
\section{Related Work}

% For time series forecasting with missing values, one line of research follows a typical grid-restricted paradigm, which first aligns irregular or incomplete sequences onto a unified time grid and then performs forecasting on the reconstructed series \cite{informer, logtrans, autoformer}. 
% In this setting, missing values are treated as defects to be removed (typically through interpolation or zero imputation), while the main effort is devoted to developing stronger forecasting architectures \cite{survey}, including RNN-based \cite{segrnn, tan2023neural}, CNN-based \cite{timesnet, micn, fact}, MLP-based \cite{dlinear, timemixer, olinear, amd,hu2026bridging}, GNN-based~\cite{timefilter,mtgnn,msgnet}, and Transformer-based \cite{informer, fedformer, timebridge, patchtst, itransformer} models.
% These methods have shown reasonable effectiveness under regular sampling or mild missingness. Their underlying assumption, however, remains unchanged: missingness is viewed only as a source of input degradation, rather than as part of the forecasting target itself \cite{informer}. Consequently, they still predict values for all future time steps and evaluate the results only on positions with available labels, which effectively treats the future as a fully predictable uniform grid and leaves future observability unmodeled.

For time series forecasting with missing values, one line of research follows a typical grid-restricted paradigm, which first aligns irregular or incomplete sequences onto a unified time grid and then performs forecasting on the reconstructed series \cite{informer, logtrans, autoformer}. In this setting, missing values are treated as defects to be removed (typically through interpolation or zero imputation), while the main effort is devoted to developing stronger forecasting architectures \cite{survey}, including RNN-based \cite{segrnn, tan2023neural}, CNN-based \cite{timesnet, micn, fact}, MLP-based \cite{dlinear, timemixer, olinear, amd,hu2026bridging}, GNN-based~\cite{timefilter,mtgnn,msgnet}, and Transformer-based \cite{informer, fedformer, timebridge, patchtst, itransformer} models. While effective under regular sampling or mild missingness, these methods retain the same underlying assumption: missingness is treated only as a source of input degradation rather than as part of the forecasting target itself \cite{informer}. As a result, they still predict values for all future time steps and evaluate only on positions with available labels, effectively treating the future as a fully predictable uniform grid and leaving future observability unmodeled \cite{hulandscape}.

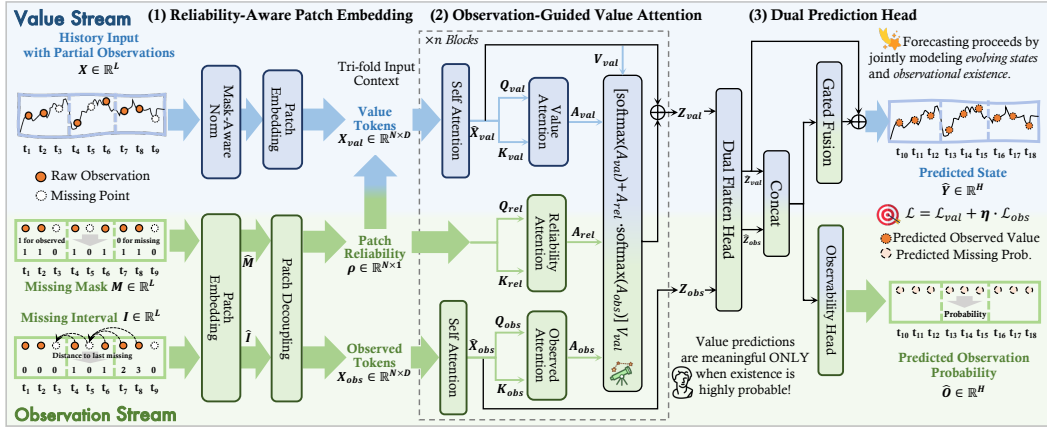
\begin{figure}[!t]
    \centering
    % Text-only overlays keep the original vector diagram intact.
    \begingroup
    \definecolor{pipelinebg}{RGB}{243,249,240}
    \definecolor{pipelinegreen}{RGB}{84,130,53}
    \resizebox{\textwidth}{!}{%
    \setlength{\unitlength}{1bp}%
    \begin{picture}(552,229)
      \put(0,0){\includegraphics[width=552bp]{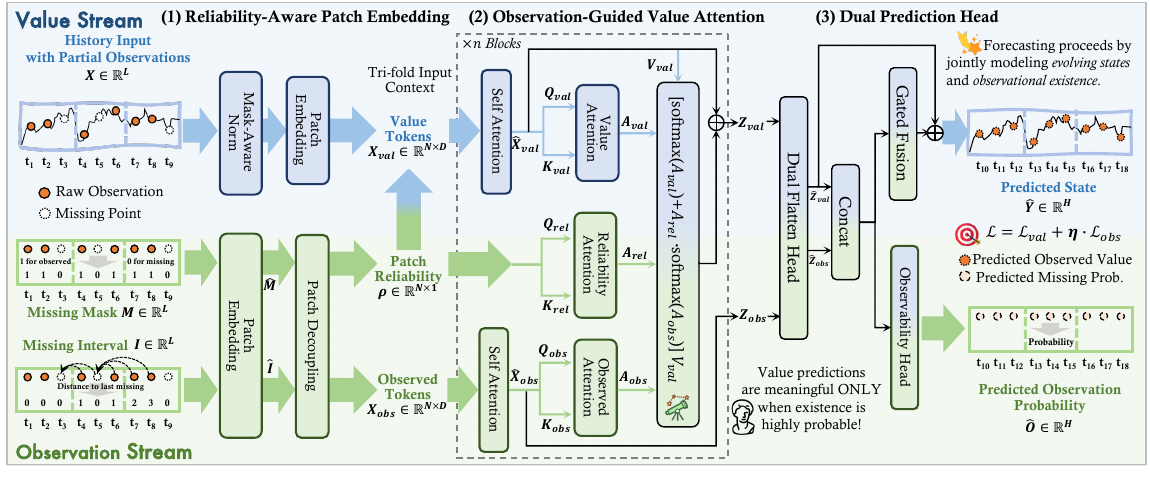}}
      \put(463,17){\color{pipelinebg}\rule{80bp}{28bp}}
      \put(463,34){\makebox(80,7){\color{pipelinegreen}\fontsize{7}{8}\selectfont\bfseries Predicted Missing Probability}}
      \put(463,22){\makebox(80,8){\fontsize{7}{8}\selectfont $\hat p_{\mathrm{miss}}=1-\hat p_{\mathrm{obs}}$}}
      \put(354,19){\color{pipelinebg}\rule{72bp}{35bp}}
      \put(354,21){\parbox[b]{72bp}{\centering\fontsize{7}{8}\selectfont Value and observation forecasts provide complementary information.}}
    \end{picture}}
    \endgroup
    \caption{The overall architecture of \NAME. (1) \textbf{Reliability-Aware Patch Embedding} refines value tokens by incorporating patch-level reliability based on observed patterns and missingness intervals. (2) \textbf{Observation-Guided Value Attention} integrates historical observation regularities into the attention mechanism to enhance value predictions. (3) \textbf{Dual Prediction Head} jointly predicts both future values and observation probabilities.}
    \label{fig:pipeline}
\end{figure}

Beyond explicit imputation, another line of research moves away from discrete reconstruction and instead models irregular sequences directly in continuous-time or weakly aligned spaces \cite{survey_irregular}, corresponding to the irregular-aware paradigm. A representative direction is based on Neural ODEs \cite{de2019gruodebayes, rubanova2019latent, schirmer2022cru, klotergens2025physiomeode}, which are naturally suited to irregular sampling because they model hidden-state evolution as a continuous-time dynamical system and can propagate states over arbitrary time intervals without requiring observations to lie on a fixed grid. Other methods build on continuous-time graph networks \cite{BiTGraph, yalavarthi2024grafiti} and combine them with patch-based alignment \cite{zhangirregular2024, luo2025hipatch}, where each patch covers the same time span but may contain a different number of valid observations, thereby relaxing strict point-wise alignment while preserving local temporal semantics. Although these methods substantially improve the modeling of historical irregularity, they still assume that the timestamps of valid future observations are known at inference time and evaluate prediction performance only on those pre-specified positions \cite{BiTGraph}. Therefore, while they move beyond the impute-then-forecast pipeline, they still do not model the generation of future observations themselves. In contrast, we explicitly treat future observability as a prediction target and jointly model whether a future observation will occur and what value it will take.

% Existing methods handle missing input values with imputation or continuous-time modeling, whereas missing future values are typically assumed to have known timestamps, with errors computed only for those valid observations. While convenient for offline evaluation, this setting overlooks a more fundamental requirement in practical deployment: before predicting a value, the model should first determine if the value will actually be observed.
\vspace{-4pt}
\section{Methodology}
\label{sec:method}

In this section, we present \NAME for jointly forecasting future values and their observability. We first formalize the problem by specifying the historical inputs, prediction targets, and probability conventions. We then introduce the dual-stream architecture and describe how its embedding, attention, and prediction modules are trained under a joint objective.

Conventional time series forecasting predicts a future sequence $Y\in\mathbb{R}^{H}$ from a historical input $X\in\mathbb{R}^{L}$, where $L$ and $H$ denote the input and prediction lengths, respectively. However, this formulation is incomplete for real-world settings with missing observations. 
Therefore, we move beyond the previous forecasting paradigm by jointly modeling future observability and future value. Specifically, in addition to the future value sequence $Y$, we introduce a future observation indicator sequence $O\in\{0,1\}^{H}$, where $O_t=1$ indicates that a valid observation exists at time step $t$, and $O_t=0$ otherwise. To infer future observability from historical missingness patterns, we further introduce two auxiliary inputs: a missing mask $M\in\{0,1\}^{L}$ and a missing interval sequence $I\in\mathbb{R}^{L}$. Here, $M_t=0$ indicates that the value at position $t$ is missing, and $M_t=1$ indicates that it is observed. The missing interval is defined as $I_t=\log(1+\delta_t)$, where $\delta_t$ denotes the distance from the current position to the most recent missing event; the logarithm is used to reduce scale variation and improve numerical stability. In summary, our model takes the triplet $(X,M,I)$ as input and jointly predicts future values and observation probabilities $(\hat{Y}, \hat{p}_{\mathrm{obs}})$. Throughout, $\hat{O}_t$ is an observation logit, $\hat{p}_{\mathrm{obs},t}=\sigma(\hat{O}_t)$ is the probability of being observed, and $\hat{p}_{\mathrm{miss},t}=1-\hat{p}_{\mathrm{obs},t}$ is the probability of missingness.

\subsection{Pipeline Overview}

As shown in \cref{fig:pipeline}, \NAME processes the inputs through two parallel streams. The value stream takes the historical value sequence $X$ and learns its temporal evolution, generating the future value prediction $\hat{Y} \in \mathbb{R}^{H}$. In parallel, the observation stream processes the missingness mask $M$ and missing intervals $I$ to capture observation regularities, producing the future observation probability $\hat{p}_{\mathrm{obs}} \in [0,1]^{H}$, where each element represents the probability of a valid observation at each time step. The entire framework consists of three key modules: (1) \textbf{Reliability-Aware Patch Embedding} divides $(X, M, I)$ into patch tokens, extracting local observation patterns and generating value tokens enhanced with patch reliability information; (2) \textbf{Observation-Guided Value Attention} models long-range dependencies within each stream and modulates the attention mechanism of value stream using the patterns learned from the observation stream, ensuring that value predictions are informed by observation regularities; (3) \textbf{Dual Prediction Head} maps the high-level representations from both streams to the future horizon, predicting both future values and observation probabilities, and incorporates observation information back into the value prediction through a gating mechanism.

\subsection{Reliability-Aware Patch Embedding}

In this stage, we first apply mask-aware normalization to $X \in \mathbb{R}^{L}$, using only the observed entries to compute the normalization statistics, which prevents distortion from missing values. We then divide $X$, $M$, and $I$ into non-overlapping patches and map them to embeddings of the same dimensionality:
\begin{equation}
    \hat{X}, \hat{M}, \hat{I} = \text{Patch\_Embedding}(X, M, I),
\end{equation}
where $\hat{X}, \hat{M}, \hat{I} \in \mathbb{R}^{N \times D}$, with $N$ denoting the number of patches and $D$ the embedding dimension.

For the observation stream, we encode missing positions and intervals using a gated fusion mechanism, which helps mitigate the challenge of directly combining features with different properties. The gated mechanism ensures that the information from missingness and intervals is appropriately modulated before integration, producing the observed token $\hat{X}_{\text{obs}} \in \mathbb{R}^{N \times D}$:
\begin{equation}
h_{\text{obs}} = \text{Linear}([\hat{M}; \hat{I}]), \quad
\hat{X}_{\text{obs}} = h_{\text{obs}} + \sigma(h_{\text{obs}}) \odot \hat{I},
\end{equation}
where $[\cdot;\cdot]$ denotes concatenation, $\sigma(\cdot)$ is the Sigmoid function, and $\odot$ is element-wise product.

Directly fusing $\hat{X}_{\text{obs}}$ into $\hat{X}$ may inject noisy cues from highly sparse patches, where limited observations provide weak and unstable evidence for value modeling. To address this, we compute a patch-level reliability score $\rho\in \mathbb{R}^{N \times 1}$ based on observation sufficiency:
\begin{equation}
\rho = \text{Linear}([\text{Mean}(\hat{M}); \text{Mean}(\hat{I})]),
\end{equation}
where $\text{Mean}(\hat{M})$ and $\text{Mean}(\hat{I})$ represent the missingness ratio and missing interval duration, respectively. The reliability score $\rho$ reflects the trustworthiness of each patch's observation information.
This score is then used to modulate the injection of observation information into the value embedding, resulting in the enhanced value token $\hat{X}_{\text{val}} \in \mathbb{R}^{N \times D}$:
\begin{equation}
    \hat{X}_{\text{val}} = \hat{X} + \text{Linear}(\rho \odot \hat{X}_{\text{obs}}).
\end{equation}
As a result, $\hat{X}_{\text{val}}$ retains the core semantics of the original value tokens while incorporating observation patterns in a reliability-aware manner.

\subsection{Observation-Guided Value Attention}

After obtaining $\hat{X}_{\text{val}}$ and $\hat{X}_{\text{obs}}$, we first apply self-attention within each stream to capture long-range dependencies across patches:
\begin{equation}
    \hat{X}_{\text{val}} = \text{Attention}(\hat{X}_{\text{val}}, \hat{X}_{\text{val}}, \hat{X}_{\text{val}}),\quad
    \hat{X}_{\text{obs}} = \text{Attention}(\hat{X}_{\text{obs}}, \hat{X}_{\text{obs}}, \hat{X}_{\text{obs}}),
\end{equation}
where the value stream $\hat{X}_{\text{val}}$ models temporal dependencies in the value sequence, while the observation stream $\hat{X}_{\text{obs}}$ captures long-range observation patterns.

However, modeling each stream independently is insufficient, as the reliability of future value predictions depends on historical observation patterns. Thus, the observation stream should not just be a supplementary branch but should influence the value stream's dependencies in a higher-level semantic space. To achieve this, we construct three attention maps: 
\begin{equation}
A_{\text{val}} = \frac{Q_{\text{val}} \cdot K_{\text{val}}^{\top}}{\sqrt{D}}, \quad
A_{\text{obs}} = \frac{Q_{\text{obs}} \cdot K_{\text{obs}}^{\top}}{\sqrt{D}}, \quad
A_{\text{rel}} = \sqrt{\rho \cdot \rho^{\top}},
\end{equation}
where $Q_{\text{val}}, K_{\text{val}} \in \mathbb{R}^{N \times D}$ are derived from $\hat{X}_{\text{val}}$, and $Q_{\text{obs}}, K_{\text{obs}} \in \mathbb{R}^{N \times D}$ are derived from $\hat{X}_{\text{obs}}$. Accordingly, $A_{\text{val}}\in\mathbb{R}^{N\times N}$ captures semantic correlations between value tokens, $A_{\text{obs}}\in\mathbb{R}^{N\times N}$ captures structural correlations between observation tokens, and $A_{\text{rel}}\in\mathbb{R}^{N\times N}$ modulates the contribution of observation patterns according to patch reliability.
We then combine these three maps into an observation-aware attention map:
\begin{equation}
A = \text{Softmax}(A_{\text{val}}) + A_{\text{rel}} \odot \text{Softmax}(A_{\text{obs}}).
\end{equation}
This fusion retains the value dependencies as the primary structure while injecting observation patterns learned from the observation stream into the value attention. Moreover, the influence of the observation stream is amplified for patches with higher reliability and suppressed for sparse patches, thus preventing over-interference with value modeling while effectively leveraging historical missingness patterns in more reliable regions.

Finally, we update the value stream using the fused attention map:
\begin{equation}
    Z_{\text{val}} = A \cdot V_{\text{val}} + \hat{X}_{\text{val}}, \quad Z_{\text{obs}} = \hat{X}_{\text{obs}},
\end{equation}
where $V_{\text{val}}$ represents the value tokens derived from $\hat{X}_{\text{val}}$. 
Semantically, $Z_{\text{val}}\in\mathbb{R}^{N\times D}$ encodes value dynamics enhanced by historical observation patterns, while $Z_{\text{obs}}\in\mathbb{R}^{N\times D}$ preserves observation information for future observability prediction.

\subsection{Dual Prediction Head}

This module takes $Z_{\text{val}}$ and $Z_{\text{obs}}$ as the input and maps these representations from both streams to the prediction horizon. Specifically, we first apply a linear Flatten Head to obtain $\hat{Z}_{\text{val}}$ and $\hat{Z}_{\text{obs}} \in \mathbb{R}^{H}$:
\begin{equation}
    \hat{Z}_{\text{val}} = \text{Flatten\_Head}(Z_{\text{val}}), \quad
    \hat{Z}_{\text{obs}} = \text{Flatten\_Head}(Z_{\text{obs}}).
\end{equation}
Based on these two representations, we first produce observation logits $\hat{O}\in \mathbb{R}^H$ and probabilities:
\begin{equation}
    \hat{O} = \text{Linear}([\hat{Z}_{\text{val}}; \hat{Z}_{\text{obs}}]),\quad \hat{p}_{\mathrm{obs}}=\sigma(\hat{O}).
\end{equation}
Both streams are used here because future observability is governed by not only historical observation patterns but also the underlying value dynamics.
The predicted observability is then used to modulate the final value prediction $\hat{Y} \in \mathbb{R}^{H}$:
\begin{equation}
    \hat{Y} = \hat{Z}_{\text{val}} + \hat{p}_{\mathrm{obs}} \odot \hat{Z}_{\text{obs}}.
\end{equation}
In this way, future observability serves as a soft gate on the contribution of observation features to value prediction, amplifying their effect on likely observed steps and attenuating it elsewhere.

\subsection{Training Loss}

The training objective consists of two components: the value prediction loss $\mathcal{L}_{\text{val}}$ and the observation prediction loss $\mathcal{L}_{\text{obs}}$. Specifically, we compute the regression loss only at positions with valid future observations and the classification loss over all prediction steps:
\begin{equation}
    \mathcal{L}_{\text{val}} = \text{MSE}(\hat{Y}, Y), \quad
    \mathcal{L}_{\text{obs}} = \text{FocalBCE}(\hat{p}_{\mathrm{obs}}, O).
\end{equation}
Here, $\mathcal{L}_{\text{val}}$ is only evaluated on time steps $O_t = 1$, while $\mathcal{L}_{\text{obs}}$ uses Focal BCE Loss to address the class imbalance between observed and missing positions. The total loss function is expressed as:
\begin{equation}
\mathcal{L} = \mathcal{L}_{\text{val}} + \eta \cdot \mathcal{L}_{\text{obs}},
\end{equation}
where $\eta$ is a hyperparameter that balances the two tasks. By jointly optimizing these two objectives, the model learns to predict both the likelihood of future observations and, under the condition of valid observations, the corresponding future state values.

% \newpage
\section{Experiments}

\subsection{\BenchName Benchmark}

\paragraph{Datasets Construction.}

% To ensure a comprehensive and realistic evaluation, \BenchName comprises 31 datasets meticulously curated from both public repositories and real-world industrial systems. 
% Specifically, we incorporate 15 public datasets from GIFT-Eval~\cite{aksu2024gift}, spanning diverse domains with granularities ranging from 5-minute to weekly intervals. 
% To further validate model performance in authentic sparse environments, we introduce 16 proprietary e-commerce datasets collected from a production online-retail platform. Each of these industrial datasets records hourly transaction volumes across different global regions. 

To ensure comprehensive and realistic evaluation, \BenchName contains 31 datasets collected from both public sources and real-world industrial systems. It includes 15 public datasets from GIFT-Eval~\cite{aksu2024gift}, covering diverse domains with frequencies ranging from 5 minutes to 1 month, and 16 proprietary e-commerce datasets from a production online-retail platform, each recording hourly transaction volumes across different global regions.

% Importantly, these proprietary datasets contain strictly \textit{naturally occurring missing values without any artificial masking}. In this context, the missingness is profoundly \textit{non-random}; it serves as an informative structural prior that encodes underlying system-level dynamics and behavioral rhythms, such as regional off-peak inactivity, holiday fluctuations, or sensor pipeline outages. Exhibiting a broad spectrum of missing ratios (ranging from less than $0.1\%$ to over $89\%$), the complete benchmark spans six distinct domains and 16 geographical regions, establishing a rigorous testbed for irregular time series forecasting. \cref{tab:datasets} summarizes the detailed statistics.

Notably, the proprietary datasets contain strictly \textit{natural missing values without artificial masking}. Their missingness is highly \textit{non-random} and encodes structural patterns, such as regional off-peak inactivity, holiday effects, and pipeline outages. With missing ratios ranging from below $0.1\%$ to above $89\%$, the full benchmark spans 6 domains and 16 geographical regions, providing a rigorous testbed for irregular time series forecasting. Table~\ref{tab:datasets} summarizes all 31 dataset-frequency units; per-split missingness statistics are provided in \appref{app:datasets}.

\begin{table}[!t]
\centering
\caption{Overview of all \BenchName datasets. Freq.: sampling frequency; \#Var: target variates; Pred Len (S/M/L): short/medium/long prediction lengths; \# Obs: total observations. BB-FS/BB-rnd abbreviate Bitbrains-Fast Storage/Bitbrains-rnd; E-com. denotes E-commerce. Missing Regimes follows Table~\ref{tab:missing_ratio_stats_long}.}
\label{tab:datasets}
\scriptsize
\setlength{\tabcolsep}{2pt}
\renewcommand{\arraystretch}{1.0}
\resizebox{\textwidth}{!}{
\begin{tabular}{llcccccccccccc}
\toprule
\BenchName &  &  &  & \multicolumn{3}{c}{\textbf{Series Length}} & & &  & \multicolumn{4}{c}{\textbf{Missing Regimes}} \\
\cmidrule(lr){5-7} \cmidrule(lr){11-14}
\textbf{Dataset} & \textbf{Domain} & \textbf{Freq.} & \textbf{Series} &
\textbf{Avg} & \textbf{Min} & \textbf{Max} & \textbf{\# Obs} & \textbf{\#Var} &
\shortstack{\textbf{Pred Len}\\\textbf{(S / M / L)}} &
\textbf{No} & \textbf{Low} & \textbf{Med} & \textbf{High} \\
\midrule
\multicolumn{14}{l}{\textit{GIFT-Eval public datasets}} \\
\midrule
\texttt{Electricity}             & Energy    & 1H    & 370    & 35,064 & 35,064 & 35,064 & 12,973,680 & 1  & [48 / 480 / 720] & \cmark & \cmark & \cmark & \cmark \\
\texttt{Electricity}             & Energy    & 1D    & 370    & 1,461  & 1,461  & 1,461  & 540,570    & 1  & [30 / -- / --]   & \cmark & \cmark & \cmark & \cmark \\
\texttt{Electricity}             & Energy    & 1W    & 370    & 208    & 208    & 208    & 76,960     & 1  & [8 / -- / --]    & \cmark & \cmark & \cmark & \cmark \\
\texttt{BB-FS}  & Cloud     & 5min  & 1,250  & 8,640  & 8,640  & 8,640  & 10,800,000 & 2  & [48 / 480 / 720] & \cmark & \cmark & \cmark & \cmark \\
\texttt{BB-FS}  & Cloud     & 1H    & 1,250  & 721    & 721    & 721    & 901,250    & 2  & [48 / 480 / 720] & \cmark & \cmark & \cmark & \cmark \\
\texttt{BB-rnd}           & Cloud     & 5min  & 500    & 8,640  & 8,640  & 8,640  & 4,320,000  & 2  & [48 / 480 / 720] & -- & \cmark & \cmark & \cmark \\
\texttt{BB-rnd}           & Cloud     & 1H    & 500    & 720    & 720    & 720    & 360,000    & 2  & [48 / -- / --]   & \cmark & \cmark & \cmark & \cmark \\
\texttt{Car Parts}               & Retail    & 1M   & 2,674  & 51     & 51     & 51     & 136,374    & 1  & [12 / -- / --]   & \cmark & -- & -- & \cmark \\
\texttt{KDD Cup 2018}            & Air Qual. & 1H   & 270    & 10,898 & 9,504  & 10,920 & 2,942,364  & 1  & [48 / 480 / 720] & -- & \cmark & \cmark & \cmark \\
\texttt{KDD Cup 2018}            & Air Qual. & 1D    & 270    & 455    & 396    & 455    & 122,791    & 1  & [30 / -- / --]   & \cmark & \cmark & \cmark & \cmark \\
\texttt{Restaurant}              & Commerce  & 1D  & 807    & 358    & 67     & 478    & 289,303    & 1  & [30 / -- / --]   & \cmark & \cmark & \cmark & \cmark \\
\texttt{Jena Weather}            & Weather   & 10min & 1      & 52,704 & 52,704 & 52,704 & 52,704     & 21 & [48 / 480 / 720] & -- & \cmark & -- & -- \\
\texttt{Jena Weather}            & Weather   & 1H    & 1      & 8,784  & 8,784  & 8,784  & 8,784      & 21 & [48 / 480 / 720] & -- & \cmark & -- & -- \\
\texttt{Hierarchical Sales}      & Retail    & 1D    & 118    & 1,825  & 1,825  & 1,825  & 215,350    & 1  & [30 / -- / --]   & -- & \cmark & -- & -- \\
\texttt{Temperature Rain}        & Weather   & 1D   & 32,072 & 725    & 725    & 725    & 23,252,200       & 1  & [30 / -- / --]   & \cmark & \cmark & \cmark & -- \\
\midrule
\multicolumn{14}{l}{\textit{Proprietary e-commerce datasets}} \\
\midrule
\texttt{ecom\_AD} & E-com. & 1H & 26 & 11,838 & 9,769 & 19,968 & 307,793 & 1 & [48 / 480 / 720] & -- & \cmark & \cmark & -- \\
\texttt{ecom\_AI} & E-com. & 1H & 9 & 11,096 & 10,910 & 12,582 & 99,862 & 1 & [48 / 480 / 720] & -- & -- & -- & \cmark \\
\texttt{ecom\_BB} & E-com. & 1H & 8 & 6,022 & 6,022 & 6,022 & 48,176 & 1 & [48 / 480 / 720] & -- & -- & \cmark & -- \\
\texttt{ecom\_BM} & E-com. & 1H & 8 & 10,667 & 9,767 & 13,366 & 85,334 & 1 & [48 / 480 / 720] & -- & -- & \cmark & -- \\
\texttt{ecom\_BY} & E-com. & 1H & 10 & 13,384 & 9,768 & 19,968 & 133,836 & 1 & [48 / 480 / 720] & -- & -- & \cmark & -- \\
\texttt{ecom\_HN} & E-com. & 1H & 16 & 7,316 & 7,316 & 7,316 & 117,056 & 1 & [48 / 480 / 720] & -- & -- & -- & \cmark \\
\texttt{ecom\_HR} & E-com. & 1H & 1 & 19,956 & 19,956 & 19,956 & 19,956 & 1 & [48 / 480 / 720] & -- & -- & \cmark & -- \\
\texttt{ecom\_JP} & E-com. & 1H & 8 & 11,953 & 11,184 & 17,333 & 95,621 & 1 & [48 / 480 / 720] & -- & \cmark & -- & -- \\
\texttt{ecom\_ME} & E-com. & 1H & 9 & 10,223 & 9,927 & 12,593 & 92,009 & 1 & [48 / 480 / 720] & -- & -- & \cmark & \cmark \\
\texttt{ecom\_NG} & E-com. & 1H & 6 & 12,920 & 12,920 & 12,920 & 77,520 & 1 & [48 / 480 / 720] & -- & -- & \cmark & -- \\
\texttt{ecom\_NO} & E-com. & 1H & 16 & 6,024 & 6,016 & 6,032 & 96,384 & 1 & [48 / 480 / 720] & -- & \cmark & \cmark & -- \\
\texttt{ecom\_NR} & E-com. & 1H & 8 & 6,032 & 6,032 & 6,032 & 48,256 & 1 & [48 / 480 / 720] & -- & \cmark & -- & -- \\
\texttt{ecom\_PL} & E-com. & 1H & 8 & 12,266 & 11,182 & 19,856 & 98,130 & 1 & [48 / 480 / 720] & -- & -- & \cmark & -- \\
\texttt{ecom\_SB} & E-com. & 1H & 24 & 12,262 & 11,141 & 19,968 & 294,287 & 1 & [48 / 480 / 720] & -- & \cmark & -- & \cmark \\
\texttt{ecom\_TA} & E-com. & 1H & 8 & 6,032 & 6,032 & 6,032 & 48,256 & 1 & [48 / 480 / 720] & -- & \cmark & -- & -- \\
\texttt{ecom\_VN} & E-com. & 1H & 16 & 11,842 & 9,769 & 13,370 & 189,474 & 1 & [48 / 480 / 720] & -- & \cmark & \cmark & -- \\
\bottomrule
\end{tabular}
}
\end{table}

\paragraph{Evaluation Protocol.}

To assess the performance of models under varying degrees of sparsity and predictive difficulty, we establish a multi-dimensional evaluation protocol along two main criteria:
\begin{itemize}[leftmargin=*,nosep]
\item \textbf{Forecasting Horizon.} Aligning with GIFT-Eval~\cite{aksu2024gift}, we evaluate models across short-, medium-, and long-term forecasting horizons, with dataset-specific prediction lengths detailed in \appref{app:datasets}.
\item \textbf{Sparsity Regime.} To decouple the impact of missingness from general forecasting complexity, we categorize the series into four distinct sparsity regimes based on their overall missing ratios: \textbf{no missing} ($0\%$), \textbf{low} ($(0,10\%]$), \textbf{medium} ($(10\%,40\%]$), and \textbf{high} ($(40\%,100\%]$).
\end{itemize}
Combining the dataset-frequency units and prediction horizons in Table~\ref{tab:datasets} with the missingness groups in Table~\ref{tab:missing_ratio_stats_long} defines a benchmark grid of 149 dataset-frequency--horizon--regime combinations. Regime labels refer to overall missingness; the missing ratios of individual splits may differ.

\begin{table}[!tb]
    \caption{Forecasting results on \BenchName benchmark. 
    For each missing-ratio regime, we first calculate the average of the test metrics across forecasting horizons for each dataset, and then report the arithmetic mean of these scores for all datasets within the same regime.
    See \cref{tab:full_missing_pred} for full results.}
    \label{tab:main_results}
    \centering
    \setlength{\tabcolsep}{1.6pt}
    \scriptsize
    \begin{threeparttable}
        \renewcommand{\arraystretch}{1.0}
        \begin{tabular}{c cccc cccc cccc ccc}
            \toprule
            \BenchName &
            \multicolumn{4}{c}{High Missing} &
            \multicolumn{4}{c}{Medium Missing} &
            \multicolumn{4}{c}{Low Missing} &
            \multicolumn{3}{c}{No Missing} \\
            \cmidrule(lr){2-5} \cmidrule(lr){6-9} \cmidrule(lr){10-13} \cmidrule(lr){14-16}
            Method & MSE$\downarrow$ & MAE$\downarrow$ & AUC$\uparrow$ & OVJE$\downarrow$
            & MSE$\downarrow$ & MAE$\downarrow$ & AUC$\uparrow$ & OVJE$\downarrow$
            & MSE$\downarrow$ & MAE$\downarrow$ & AUC$\uparrow$ & OVJE$\downarrow$
            & MSE$\downarrow$ & MAE$\downarrow$  & OVJE$\downarrow$ \\
            \midrule
            FEDformer    & 1.666 & 0.429 & 0.614 & 1.307 & 0.648 & 0.395 & 0.558 & 1.277 & 1.490 & 0.438 & 0.525 & 1.136 & 0.420 & 0.314 & 0.862 \\
            PatchTST     & 1.687 & 0.350 & 0.719 & 0.872 & 0.634 & 0.360 & 0.614 & 1.215 & 1.439 & 0.380 & 0.541 & \underline{0.924} & 0.340 & 0.255 & 0.708 \\
            Crossformer  & 1.852 & 0.369 & 0.654 & 0.944 & 0.655 & 0.380 & 0.553 & 1.243 & 1.458 & 0.400 & 0.497 & 0.990 & 0.436 & 0.287 & 0.780 \\
            iTransformer & \underline{1.590} & 0.354 & 0.680 & 0.846 & 0.638 & 0.367 & 0.566 & 1.263 & 1.452 & 0.387 & 0.524 & 1.017 & 0.324 & 0.251 & 0.730 \\
            DLinear      & 1.714 & 0.383 & 0.638 & 1.069 & 0.611 & 0.367 & 0.528 & 1.328 & \underline{1.433} & 0.387 & 0.507 & 1.245 & 0.371 & 0.286 & 1.020 \\
            TimeMixer    & 4.490 & 0.730 & 0.706 & 1.118 & 0.638 & 0.372 & 0.621 & 1.344 & 1.485 & 0.400 & 0.546 & 1.021 & 0.410 & 0.289 & 0.741 \\
            OLinear      & 1.622 & \underline{0.328} & \underline{0.740} & \underline{0.762} & \underline{0.608} & 0.352 & 0.623 & 1.475 & 1.438 & \underline{0.376} & 0.525 & 0.958 & 0.317 & 0.247 & 0.695 \\
            WPMixer      & 1.601 & 0.342 & 0.578 & 0.930 & 0.615 & \underline{0.349} & 0.536 & 1.202 & 1.442 & 0.377 & 0.500 & 1.255 & 0.331 & 0.253 & 0.947 \\
            MICN         & 8.191 & 0.888 & 0.676 & 1.472 & 7.253 & 1.190 & \underline{0.657} & 1.468 & 13.961 & 1.399 & \underline{0.561} & 1.194 & 17.814 & 1.685 & 0.984 \\
            FACT         & 1.644 & 0.334 & 0.730 & 0.823 & 0.617 & 0.352 & 0.581 & \underline{1.176} & 1.461 & 0.383 & 0.551 & 0.927 & \underline{0.307} & \underline{0.240} & \underline{0.639} \\
            \NAME (Ours)      & \textbf{1.546} & \textbf{0.303} & \textbf{0.805} & \textbf{0.611} & \textbf{0.555} & \textbf{0.285} & \textbf{0.704} & \textbf{0.914} & \textbf{1.362} & \textbf{0.349} & \textbf{0.588} & \textbf{0.762} & \textbf{0.296} & \textbf{0.229} & \textbf{0.477} \\
            \bottomrule
        \end{tabular}
    \end{threeparttable}
\end{table}

\vspace{-7pt}

\begin{table}[!tb]
    % \caption{Value forecasting performance across missingness regimes. Results compare MSE and MAE without the missingness classification head. Values represent the two-stage arithmetic mean, calculated first across forecasting horizons per dataset, and then across all datasets within each regime.
    % See \cref{tab:value_prediction_full} for full results.}
    \caption{Value forecasting performance across missingness regimes without the missingness classification head. Results are reported as two-stage arithmetic means, aggregated first over forecasting horizons within each dataset and then over datasets within each regime. See \cref{tab:value_prediction_full} for full results.}
    \label{tab:value_forecast_results}
    \centering
    \setlength{\tabcolsep}{10.5pt}
    \scriptsize
    \begin{threeparttable}
        \renewcommand{\arraystretch}{1.0}
        \begin{tabular}{c cc cc cc cc}
            \toprule
            \BenchName &
            \multicolumn{2}{c}{High Missing} &
            \multicolumn{2}{c}{Medium Missing} &
            \multicolumn{2}{c}{Low Missing} &
            \multicolumn{2}{c}{No Missing} \\
            \cmidrule(lr){2-3} \cmidrule(lr){4-5} \cmidrule(lr){6-7} \cmidrule(lr){8-9}
            Method 
            & MSE$\downarrow$ & MAE$\downarrow$
            & MSE$\downarrow$ & MAE$\downarrow$
            & MSE$\downarrow$ & MAE$\downarrow$
            & MSE$\downarrow$ & MAE$\downarrow$ \\
            \midrule
            FEDformer    & 1.688 & 0.408 & 0.657 & 0.384 & 1.496 & 0.443 & 0.423 & 0.316 \\
            PatchTST     & 1.766 & 0.337 & 0.650 & 0.354 & 1.446 & 0.372 & 0.351 & 0.259 \\
            Crossformer  & 1.909 & 0.369 & 0.659 & 0.368 & 1.587 & 0.440 & 0.434 & 0.291 \\
            iTransformer & 1.611 & 0.348 & 0.634 & 0.353 & 1.462 & 0.379 & 0.326 & 0.251 \\
            DLinear      & 1.716 & 0.369 & 0.614 & 0.354 & 1.438 & 0.380 & 0.367 & 0.283 \\
            TimeMixer    & 1.935 & 0.452 & 0.636 & 0.354 & 1.456 & 0.381 & 0.396 & 0.283 \\
            OLinear      & \underline{1.586} & \underline{0.311} & \underline{0.596} & \underline{0.333} & \underline{1.427} & \underline{0.364} & 0.301 & 0.240 \\
            WPMixer      & 1.633 & 0.327 & 0.637 & 0.345 & 1.434 & 0.369 & 0.303 & 0.241 \\
            MICN         & 9.151 & 0.925 & 88.328 & 2.585 & 16.245 & 1.611 & 19.409 & 1.737 \\
            FACT         & 1.683 & 0.340 & 0.627 & 0.348 & 1.447 & 0.371 & \underline{0.293} & \underline{0.234} \\
            \NAME (Ours) & \textbf{1.501} & \textbf{0.294} & \textbf{0.557} & \textbf{0.276} & \textbf{1.403} & \textbf{0.355} & \textbf{0.284} & \textbf{0.226} \\
            \bottomrule
        \end{tabular}
    \end{threeparttable}
    \vspace{-10pt}
\end{table}

\subsection{Experimental Setup}

\paragraph{Baselines.}
We compare \NAME against a comprehensive set of baselines,  including Transformer-based methods: PatchTST~\cite{patchtst}, iTransformer~\cite{itransformer}, Crossformer~\cite{crossformer} and FEDformer~\cite{fedformer}; Linear-based methods: OLinear~\cite{olinear}, WPMixer~\cite{wpmixer}, TimeMixer~\cite{timemixer} and DLinear~\cite{dlinear}; CNN-based methods: FACT~\cite{fact} and MICN~\cite{micn}.
All baselines are adapted to the missing-aware setting by receiving \textbf{\textit{zero-filled}} inputs with accompanying observation masks.

% \paragraph{Implementation details.}
% All experiments are implemented in PyTorch~\cite{pytorch} and run on a single NVIDIA B200 180GB GPU. We use Adam~\cite{adam} with the learning rate chosen from $\{1e^{-3}, 1e^{-4}, 5e^{-5}\}$, and set $\eta$ for $\mathcal{L}_{\text{val}}$ to $1.0$. More hyperparameter details are provided in \appref{app:hyperparams}.

\paragraph{Metrics.}
% To evaluate the dual-stream architecture, we employ task-specific metrics and propose a novel joint evaluation criterion.
% For the observation stream, we adopt \textbf{AUC} to robustly measure the classification of imbalanced missing events.
% For the value stream, we utilize \textbf{MSE} and \textbf{MAE}, computed exclusively on ground-truth observed timestamps to ensure fair regression assessment.
% Furthermore, guided by our ``existence precedes value'' philosophy, numerical predictions are meaningful only when accurately anticipated. Thus, we propose the Observation-Value Joint Entropy (\textbf{OVJE}) metric to comprehensively quantify this coupled capability. 

To evaluate the dual-stream pipeline, we report task-specific metrics and a joint criterion. For the observation stream, \textbf{ROC-AUC} measures discrimination, with observed positions ($O_t=1$) as the positive class; it is undefined for single-class targets. For the value stream, \textbf{MSE} and \textbf{MAE} are computed only on ground-truth observed positions. We additionally report the Observation-Value Joint Entropy (\textbf{OVJE}) as a sample-aligned score of both outputs.

Let $o_t=O_t\in\{0,1\}$ and $\hat p_t=\hat p_{\mathrm{obs},t}$ denote the target observation indicator and its predicted probability. The implementation uses a symmetric relative error (sMAPE, without a factor of 100):
\begin{equation}
e_t=\frac{2|y_t-\hat y_t|}{|y_t|+|\hat y_t|+\epsilon},\qquad q_t=\exp(-e_t),\qquad \epsilon=10^{-8}.
\label{eq:ovje_quality}
\end{equation}
Here $q_t$ is a value-quality factor, not a normalized predictive density over $y_t$. We clip $\hat p_t$ to $[\epsilon,1-\epsilon]$ before taking logarithms. The joint score is
\begin{align}
\mathrm{OVJE}
&=-\frac{1}{T}\sum_{t=1}^{T}\left[o_t\log(\hat p_tq_t)+(1-o_t)\log(1-\hat p_t)\right] \label{eq:ovje}\\
&=\underbrace{-\frac{1}{T}\sum_{t=1}^{T}\left[o_t\log\hat p_t+(1-o_t)\log(1-\hat p_t)\right]}_{\text{Bernoulli NLL}}
\ \ \ \ \ \qquad+\underbrace{\frac{1}{T}\sum_{t:o_t=1}e_t}_{\text{observed-position sMAPE contribution}}. \nonumber
\end{align}
Both terms are averaged over all evaluated positions, not only observed targets. Missing targets contribute only observation NLL. Lower OVJE is better: it complements separate metrics through a sample-aligned, confidence-sensitive score, while decomposing exactly into the two terms above. An example with equal AUC/MAE but different OVJE appears in \appref{app:metric}.

\begin{table}[!tb]
    \caption{Ablation studies of \NAME. Within each missing-ratio regime, performance is summarized using the same two-stage arithmetic averaging procedure as in \cref{tab:main_results}. } 
    \label{tab:ablation}
    \centering
    \scriptsize
    \setlength{\tabcolsep}{3.4pt}
    \scriptsize
    \begin{threeparttable}
        \renewcommand{\arraystretch}{1.0}
        \begin{tabular}{c cccc cccc cccc}
            \toprule
            \BenchName &
            \multicolumn{4}{c}{High Missing} &
            \multicolumn{4}{c}{Medium Missing} &
            \multicolumn{4}{c}{Low Missing} \\
            % \multicolumn{4}{c}{No Missing} \\
            \cmidrule(lr){2-5} \cmidrule(lr){6-9} \cmidrule(lr){10-13} 
            % \cmidrule(lr){14-17}
            Variant & MSE$\downarrow$ & MAE$\downarrow$ & AUC$\uparrow$ & OVJE$\downarrow$
            & MSE$\downarrow$ & MAE$\downarrow$ & AUC$\uparrow$ & OVJE$\downarrow$
            & MSE$\downarrow$ & MAE$\downarrow$ & AUC$\uparrow$ & OVJE$\downarrow$ \\
            % & MSE$\downarrow$ & MAE$\downarrow$ & AUC$\uparrow$ & OVJE$\downarrow$ \\
            \midrule
            w/o Missing Interval     & 1.580 & 0.321 & 0.744 & 0.733 & 0.586 & 0.289 & 0.663 & 0.977 & 1.522 & \underline{0.372} & 0.544 & 0.836 \\ % & 0.296 & 0.229 & - & 0.477 \\
            w/o Focal Loss  & 1.509 & 0.322 & \underline{0.760} & \underline{0.691} & 0.584 & 0.291 & \underline{0.686} & 1.042 & 1.522 & 0.375 & \underline{0.557} & \underline{0.777} \\ % & 0.300 & 0.216 & - & 0.463 \\
            w/o Mask-Aware Norm & 1.549 & 0.367 & 0.745 & 0.731 & 0.617 & 0.333 & 0.684 & 0.945 & \underline{1.517} & 0.379 & 0.549 & 0.801 \\ % & 0.296 & 0.229 & - & 0.477 \\
            w/o Obs-Conditioned Attn      & 1.538 & 0.326 & 0.732 & 0.724 & 0.585 & 0.287 & 0.673 & \underline{0.943} & 1.518 & 0.374 & 0.530 & 0.799 \\ % & 0.314 & 0.242 & - & 0.492 \\
            w/o Patch Reliability     & \underline{1.500} & 0.321 & 0.751 & 0.708 & \underline{0.583} & \underline{0.286} & 0.683 & 0.958 & 1.522 & 0.377 & 0.553 & 0.801 \\ % & 0.309 & 0.236 & - & 0.481 \\
            w/o Residual Fusion      & \textbf{1.491} & \underline{0.318} & 0.751 & 0.697 & 0.585 & \textbf{0.285} & 0.672 & 0.967 & 1.526 & \underline{0.372} & 0.542 & 0.808 \\ % & 0.291 & 0.212 & - & 0.471 \\
            \NAME (Full)    & 1.546 & \textbf{0.303} & \textbf{0.805} & \textbf{0.611} & \textbf{0.555} & \textbf{0.285} & \textbf{0.704} & \textbf{0.914} & \textbf{1.362} & \textbf{0.349} & \textbf{0.588} & \textbf{0.762} \\ % 0.296 & 0.229 & - & 0.477 \\
            \bottomrule
        \end{tabular}
    \end{threeparttable}
\end{table}

\begin{figure}[!t]
    \centering
    \includegraphics[width=\textwidth]{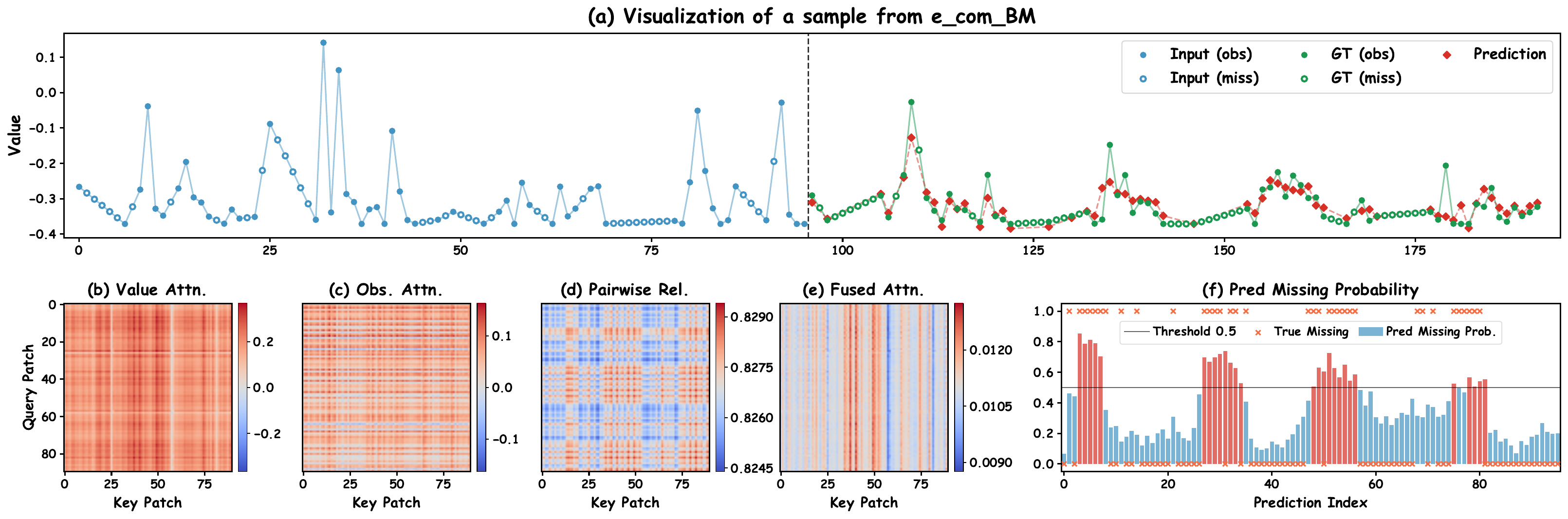}
    \vspace{-8pt}
    % \caption{Mechanistic visualization of \NAME on a medium irregular sample from \texttt{ecom\_BM}. 
    % (a) \textbf{Forecasting trajectory}: Filled and hollow circles denote valid and missing observations, respectively. Despite historical sparsity, the model robustly extrapolates the ground truth (green) via its predictions (red). 
    % (b)--(e) \textbf{Attention dynamics}: Step-by-step formation of the observation-conditioned routing mechanism, including (b) \textbf{value attention}, (c) \textbf{observed attention}, (d) \textbf{reliability attention}, and (e) \textbf{fused attention}. 
    % (f) \textbf{Existence inference}: Predicted missingness probability across the forecast horizon. Red crosses mark actual missing events, and the horizontal line indicates the 0.5 threshold, validating the model's precise anticipation of data existence.}
    \caption{Visualization of \NAME on a medium-irregular sample from \texttt{ecom\_BM}. 
    (a) \textbf{Forecasting trajectory}: Filled and hollow circles denote valid and missing observations. Despite historical sparsity, the model closely follows the ground truth (green) with predictions (red). 
    (b)--(e) \textbf{Attention dynamics}: Construction of the observation-conditioned routing mechanism, including (b) value attention, (c) observed attention, (d) reliability attention, and (e) fused attention. 
    (f) \textbf{Existence inference}: The complementary missingness probability $\hat{p}_{\mathrm{miss}}=1-\hat{p}_{\mathrm{obs}}$ over the forecast horizon. \textcolor{red}{$\times$} denotes actual missing events, and the horizontal line denotes the 0.5 threshold, indicating accurate future observability prediction.}
    \label{fig:model_visualization}
    \vspace{-8pt}
\end{figure}

\subsection{Main Results}

To distinguish joint prediction from value-only forecasting, we evaluate \NAME under two settings: joint forecasting (with the classification head) and pure value regression (without it).

\paragraph{Joint Forecasting of Observability and Value.} 
In \cref{tab:main_results}, \NAME achieves the lowest OVJE in every sparsity regime. Averaging OVJE equally across the four regimes, it improves over the strongest baseline, FACT~\cite{fact}, by approximately \textbf{\textit{22.5\%}}. 
The aggregate joint-score gains are particularly pronounced in sparse regimes. They support predicting observability as an additional output, but do not establish that future-observability supervision universally improves MAE or MSE. Its incremental value benefit depends on the observation mechanism, as examined by the controlled experiments in \appref{app:controlled}.

\paragraph{Historical Missing Patterns as Vital Context.} 
To verify our claim that \textit{historical missing patterns provide vital context for understanding the system's underlying dynamics}, \cref{tab:value_forecast_results} evaluates \NAME tasked solely with value regression (without the classification head). 
Even without explicit observability supervision, \NAME consistently yields the lowest MSE and MAE. The improvements peak under medium missingness (\textbf{\textit{17.1\%}} MAE vs. OLinear), where irregular distribution shifts heavily disrupt standard attention mechanisms. This supports the use of historical observation information. Under IID missingness, such information identifies valid inputs; learning state-related observation dynamics requires a structured, state-relevant observation process.

\vspace{-5pt}
\subsection{Ablation Studies}
To validate the effectiveness of \NAME, we evaluate several variants in \cref{tab:ablation}. The results characterize each component's contribution, whose magnitude and direction can vary across metrics and missingness regimes.
% \textbf{Mask-Aware Normalization:} Removing this component causes severe performance drops, particularly in the High Missing regime. This indicates that filtering out NaN-filled zeros is essential to prevent skewed statistical distributions during normalization.
% \textbf{Obs-Conditioned Attn \& Patch Reliability:} Without these modules, the model fails to adapt its attention to observation quality. This confirms that explicitly differentiating informative patches and routing attention based on missingness priors is critical for sparse representation learning.
% \textbf{Missing Interval:} Removing the time-since-last-observation features degrades predictive accuracy, validating its role in capturing the irregular temporal gaps between valid signals.
% \textbf{Focal Loss:} Replacing the focal loss directly diminishes the existence inference capability (AUC), supporting its use for imbalanced observation targets, without implying uniformly better probability calibration.
% \textbf{Residual Fusion:} Simplifying the final gated fusion hampers the model, verifying the value of observation-aware decoding before outputting the final predictions.
\begin{itemize}[leftmargin=*,nosep]
\item \textbf{Mask-Aware Normalization:} Removing this component causes the largest drop, especially under High Missing, showing that excluding NaN-filled zeros is crucial for stable normalization.
\item \textbf{Obs-Conditioned Attn \& Patch Reliability:} Without these modules, the model fails to adapt its attention to observation quality. This confirms that explicitly differentiating informative patches and routing attention based on missingness priors is critical for sparse representation learning.
\item \textbf{Missing Interval:} Removing the time-since-last-observation features degrades predictive accuracy, validating its role in capturing the irregular temporal gaps between valid signals.
\item \textbf{Focal Loss:} Replacing the focal loss diminishes the existence inference capability (AUC), supporting its use for imbalanced observation targets, without implying better probability calibration.
\item \textbf{Residual Fusion:} Removing this component worsens MAE, AUC, and OVJE under High Missing, but improves MSE from 1.546 to 1.491, indicating a metric-dependent contribution.
\end{itemize}
\vspace{-5pt}

\paragraph{Controlled missingness and contribution analysis.}
On five complete datasets, IID masks yield near-chance observation AUC; BLOCK-50 reaches $0.930$, but does not improve value accuracy over the matched backbone. Under SHIFT-10-50, a simpler mask-aware baseline performs better. M/V0--V3 controls attribute No-Missing gains to architecture and auxiliary optimization, and IID gains to input-validity awareness. Experimental settings and both result tables appear in \appref{app:controlled}.

\subsection{Model Analysis}

\begin{figure}[!t]
    \centering
    \includegraphics[width=\textwidth]{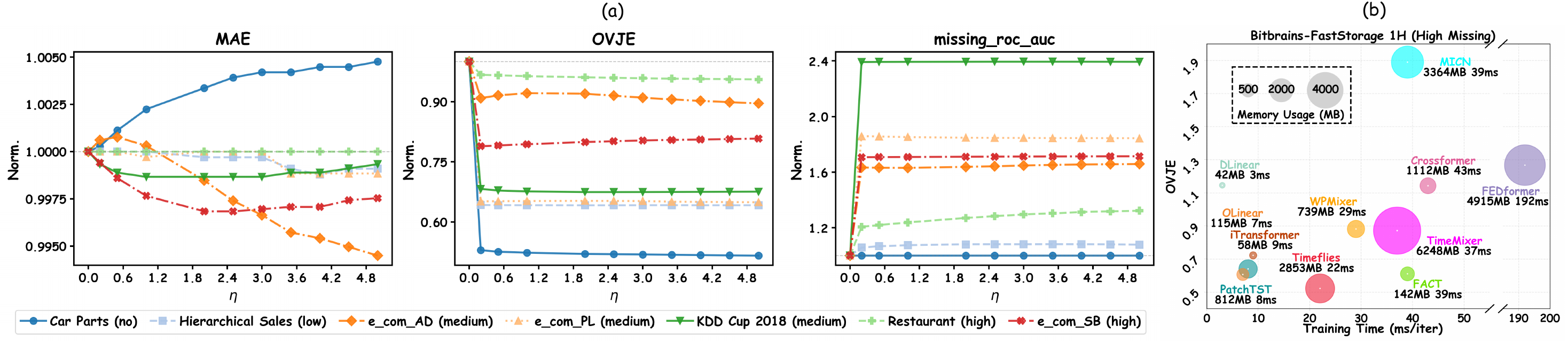}
    \vspace{-12pt}
    \caption{(a) Sensitivity of \NAME to the missingness loss weight $\eta$ across datasets spanning four missing-ratio regimes. Parenthesized labels denote the missingness regime. (b) Model efficiency comparison under Bitbrains-FastStorage 1H dataset with high missing.}
    \vspace{-8pt}
    \label{fig:loss_weight_sensitivity}
\end{figure}

\paragraph{Case study on Observation-Guided Value Attention.}
% To analyze how \NAME processes irregular missingness, we visualize its internal attention dynamics on a real-world sample. As shown in \cref{fig:model_visualization}(a), despite extreme discontinuity in the historical context window, the model accurately captures and extrapolates the underlying state evolution. This robust performance underscores the framework's capability to distill coherent temporal structures from heavily fragmented inputs.
Figure~\ref{fig:model_visualization}(a)--(e) illustrates trend recovery from fragmented history and attention that emphasizes informative patches. The detailed interpretation is retained in \appref{app:attention_case}. Panel (f) shows missingness probability $\hat{p}_{\mathrm{miss}}=1-\hat{p}_{\mathrm{obs}}$; higher plotted values indicate future missing events. Training and OVJE instead use $\hat{p}_{\mathrm{obs}}=\sigma(\hat{O})$, with observed positions as the positive class.

\paragraph{Influence of Loss Weight $\eta$} 
As shown in \cref{fig:loss_weight_sensitivity}(a), sensitivity to the loss weight $\eta$ depends strongly on dataset missingness. In the fully observed setting (Car Parts Dataset), MAE is nearly unaffected by $\eta$. Under moderate-to-high missingness, the effect on MAE is dataset-dependent: some datasets improve as $\eta$ increases, while others change little or degrade. The improvements in OVJE and ROC-AUC on several datasets when $\eta > 0$ support explicit mask supervision. ROC-AUC is unchanged when both observation labels and probabilities are complemented, as in the missingness visualization. The range $\eta \in [1.0, 3.0]$ can serve as a starting point for dataset-specific tuning.

\paragraph{Efficiency Analysis.}
We comprehensively compare the forecasting performance (OVJE), training speed, and memory footprint of \NAME and other baselines, using the official model configurations and the same batch size. As shown in \cref{fig:loss_weight_sensitivity}(b), the efficiency of \NAME exceeds other Transformer-based and CNN-based methods.
% 画图，一个单变量一个多变量

\vspace{-6pt}

\section{Conclusion}
In this work, we challenge the conventional \textbf{\textit{perfect foresight}} assumption in time series forecasting, arguing that predicting \textit{whether} an observation occurs is as critical as predicting its value. 
We propose \textbf{\NAME} to jointly model observational existence and state evolution, the \textbf{\BenchName} benchmark with natural missingness, and the Observation-Value Joint Entropy (OVJE) score to evaluate both outputs. 
Experiments show strong aggregate joint performance and competitive value forecasting. Additional value gains are data- and mechanism-dependent: predictable observability alone is insufficient when it contains no state-relevant information. 
Overall, by bridging observability and numerical regression, we open a new avenue for time series forecasting in irregular settings.

\clearpage
\bibliographystyle{unsrt}
\bibliography{neurips_2026}

%%%%%%%%%%%%%%%%%%%%%%%%%%%%%%%%%%%%%%%%%%%%%%%%%%%%%%%%%%%%
\newpage
\appendix

\section{Additional Experimental Details}
\label{app:exp_details}

\subsection{Dataset Details}
\label{app:datasets}
\cref{tab:datasets} provides a detailed characterization of the \BenchName benchmark, which integrates 15 public datasets from \textit{GIFT-Eval} with 16 proprietary e-commerce datasets spanning domains such as energy, cloud computing, and retail. For each dataset, we report key temporal statistics---including sampling frequency, series count, and sequence length distributions---alongside the number of target variates and multi-horizon prediction lengths ($S/M/L$).

To evaluate model robustness under varying data sparsity, we stratify the series into four missingness regimes: \textit{None} (0\%), \textit{Low} ($(0,10\%]$), \textit{Medium} ($(10\%,40\%]$), and \textit{High} ($(40\%,100\%]$). While public datasets exhibit heterogeneous patterns---with some, like \textit{Electricity}, spanning all regimes and others localized to specific subsets---the proprietary datasets are predominantly hourly and characterized by higher missing ratios. This distribution reflects the inherent sparsity of real-world transactional data.

Collectively, \BenchName offers a broad spectrum of missingness severities, facilitating a rigorous evaluation of forecasting models across controlled and realistic scenarios.

% The dataset overview table is now in Sec. 4.1.

Table~\ref{tab:missing_ratio_stats_long} details the per-split missingness ratios across all 
dataset--frequency--regime configurations under the GIFT-Eval 80/10/10 protocol. Car Parts in the High regime has no observed validation or test targets; observed-target MAE/MSE are therefore undefined for this group. 
While most datasets exhibit ``front-loaded'' missingness (higher sparsity in training 
than in evaluation), the e-commerce datasets (\texttt{ecom\_NG} and \texttt{ecom\_VN}) 
present a more challenging inverse pattern. Specifically, in the Medium regime, 
\texttt{ecom\_NG}'s missing ratio escalates from $0.15\%$ in training to $94.43\%$ 
in testing; \texttt{ecom\_VN} follows a similar trajectory, rising from $1.13\%$ to 
$93.03\%$. High missingness is already present during validation ($54.33\%$ for \texttt{ecom\_NG} and $83.48\%$ for \texttt{ecom\_VN}). These natural regime changes therefore differ from the previously unseen test-only shift in \appref{app:controlled}; they are stress tests rather than general robustness guarantees.

{\scriptsize
\begin{longtable}{lllcccc}
\caption{Missing-ratio statistics for each dataset under different frequencies and missing regimes. We report the overall missing ratio, together with the missing ratios in the training, validation, and test splits. Missing regimes are abbreviated as \textbf{High}, \textbf{Medium}, \textbf{Low}, and \textbf{No}.}
\label{tab:missing_ratio_stats_long} \\

\toprule
\textbf{Dataset} & \textbf{Freq.} & \textbf{Missing Ratio} & \textbf{Overall} & \textbf{Train} & \textbf{Val} & \textbf{Test} \\
\midrule
\endfirsthead

\multicolumn{7}{c}{\tablename\ \thetable{} -- continued from previous page} \\
\toprule
\textbf{Dataset} & \textbf{Freq.} & \textbf{Missing Ratio} & \textbf{Overall} & \textbf{Train} & \textbf{Val} & \textbf{Test} \\
\midrule
\endhead

\midrule
\multicolumn{7}{r}{Continued on next page} \\
\endfoot

\bottomrule
\endlastfoot

\multicolumn{7}{l}{\textit{GIFT-Eval public datasets}} \\
\midrule

\texttt{Bitbrains-Fast Storage} & 5T & High   & 77.14\% & 83.57\% & 54.60\% & 48.25\% \\
                                &    & Low    & 0.26\%  & 0.16\%  & 1.00\%  & 0.27\% \\
                                &    & Medium & 19.51\% & 22.74\% & 7.25\%  & 5.89\% \\
                                &    & No     & 0.00\%  & 0.00\%  & 0.00\%  & 0.00\% \\
                                & H  & High   & 77.00\% & 83.50\% & 54.12\% & 48.26\% \\
                                &    & Low    & 3.81\%  & 2.70\%  & 0.00\%  & 16.35\% \\
                                &    & Medium & 19.25\% & 22.62\% & 5.88\%  & 5.88\% \\
                                &    & No     & 0.00\%  & 0.00\%  & 0.00\%  & 0.00\% \\
\midrule

\texttt{Bitbrains-rnd} & 5T & High   & 69.45\% & 81.75\% & 26.21\% & 14.29\% \\
                       &    & Low    & 4.15\%  & 5.15\%  & 0.30\%  & 0.03\% \\
                       &    & Medium & 33.98\% & 42.44\% & 0.23\%  & 0.02\% \\
                       & H  & High   & 69.68\% & 82.04\% & 26.15\% & 14.32\% \\
                       &    & Low    & 4.13\%  & 5.16\%  & 0.00\%  & 0.00\% \\
                       &    & Medium & 34.52\% & 43.15\% & 0.00\%  & 0.00\% \\
                       &    & No     & 0.00\%  & 0.00\%  & 0.00\%  & 0.00\% \\
\midrule

\texttt{Car Parts} & M & High & 72.75\% & 65.26\% & 100.00\% & 100.00\% \\
                   &   & No   & 0.00\%  & 0.00\%  & 0.00\%   & 0.00\% \\
\midrule

\texttt{Electricity} & D  & High   & 67.03\% & 83.59\% & 2.05\% & 0.00\% \\
                     &    & Low    & 5.07\%  & 6.34\%  & 0.00\% & 0.00\% \\
                     &    & Medium & 25.44\% & 31.82\% & 0.00\% & 0.00\% \\
                     &    & No     & 0.00\%  & 0.00\%  & 0.00\% & 0.00\% \\
                     & H  & High   & 67.04\% & 83.55\% & 2.04\% & 0.00\% \\
                     &    & Low    & 5.08\%  & 6.35\%  & 0.00\% & 0.00\% \\
                     &    & Medium & 25.44\% & 31.80\% & 0.00\% & 0.00\% \\
                     &    & No     & 0.00\%  & 0.00\%  & 0.00\% & 0.00\% \\
                     & 1W & High   & 66.99\% & 83.68\% & 2.14\% & 0.00\% \\
                     &    & Low    & 4.81\%  & 6.02\%  & 0.00\% & 0.00\% \\
                     &    & Medium & 25.45\% & 31.89\% & 0.00\% & 0.00\% \\
                     &    & No     & 0.00\%  & 0.00\%  & 0.00\% & 0.00\% \\
\midrule

\texttt{Hierarchical Sales} & D & Low & 1.48\% & 1.16\% & 2.75\% & 2.73\% \\
\midrule

\texttt{Jena Weather} & 10T & Low & 0.02\% & 0.02\% & 0.00\% & 0.00\% \\
                      & H   & Low & 0.01\% & 0.01\% & 0.00\% & 0.00\% \\
\midrule

\texttt{KDD Cup 2018} & D & High   & 64.79\% & 60.93\% & 80.00\% & 80.43\% \\
                      &   & Low    & 4.59\%  & 5.50\%  & 0.67\%  & 1.15\% \\
                      &   & Medium & 19.48\% & 19.80\% & 16.44\% & 19.93\% \\
                      &   & No     & 0.00\%  & 0.00\%  & 0.00\%  & 0.00\% \\
                      & H & High   & 56.68\% & 53.35\% & 67.20\% & 72.83\% \\
                      &   & Low    & 6.91\%  & 7.87\%  & 3.14\%  & 3.03\% \\
                      &   & Medium & 18.18\% & 20.42\% & 7.93\%  & 10.45\% \\
\midrule

\texttt{Restaurant} & D & High   & 44.19\% & 45.08\% & 43.07\% & 38.35\% \\
                    &   & Low    & 4.03\%  & 4.27\%  & 2.91\%  & 3.24\% \\
                    &   & Medium & 19.50\% & 20.05\% & 17.71\% & 17.02\% \\
                    &   & No     & 0.00\%  & 0.00\%  & 0.00\%  & 0.00\% \\
\midrule

\texttt{Temperature Rain} & D & Low    & 2.67\%  & 3.32\%  & 0.04\% & 0.03\% \\
                          &   & Medium & 17.74\% & 22.17\% & 0.00\% & 0.00\% \\
                          &   & No     & 0.00\%  & 0.00\%  & 0.00\% & 0.00\% \\

\midrule
\multicolumn{7}{l}{\textit{Proprietary e-commerce datasets}} \\
\midrule

\texttt{ecom\_AD} & 1H & Low    & 0.10\%  & 0.09\%  & 0.00\%  & 0.29\% \\
                  &    & Medium & 17.32\% & 10.85\% & 27.79\% & 58.53\% \\
\midrule

\texttt{ecom\_AI} & 1H & High & 56.01\% & 58.55\% & 30.08\% & 61.62\% \\
\midrule

\texttt{ecom\_BB} & 1H & Medium & 22.52\% & 20.24\% & 34.05\% & 29.19\% \\
\midrule

\texttt{ecom\_BM} & 1H & Medium & 27.21\% & 22.34\% & 47.84\% & 45.48\% \\
\midrule

\texttt{ecom\_BY} & 1H & Medium & 24.43\% & 16.52\% & 49.27\% & 62.91\% \\
\midrule

\texttt{ecom\_HN} & 1H & High & 83.50\% & 86.74\% & 72.23\% & 68.83\% \\
\midrule

\texttt{ecom\_HR} & 1H & Medium & 24.40\% & 30.50\% & 0.00\% & 0.00\% \\
\midrule

\texttt{ecom\_JP} & 1H & Low & 0.24\% & 0.30\% & 0.00\% & 0.00\% \\
\midrule

\texttt{ecom\_ME} & 1H & High   & 63.33\% & 67.23\% & 48.69\% & 46.78\% \\
                  &    & Medium & 24.40\% & 29.63\% & 4.53\%  & 2.46\% \\
\midrule

\texttt{ecom\_NG} & 1H & Medium & 14.99\% & 0.15\% & 54.33\% & 94.43\% \\
\midrule

\texttt{ecom\_NO} & 1H & Low    & 5.40\%  & 6.42\%  & 1.16\%  & 1.49\% \\
                  &    & Medium & 36.32\% & 39.34\% & 25.62\% & 22.89\% \\
\midrule

\texttt{ecom\_NR} & 1H & Low & 6.68\% & 5.76\% & 9.95\% & 10.76\% \\
\midrule

\texttt{ecom\_PL} & 1H & Medium & 17.02\% & 19.39\% & 11.06\% & 3.99\% \\
\midrule

\texttt{ecom\_SB} & 1H & High & 79.16\% & 81.14\% & 73.67\% & 68.82\% \\
                  &    & Low  & 0.08\%  & 0.10\%  & 0.00\%  & 0.00\% \\
\midrule

\texttt{ecom\_TA} & 1H & Low & 0.13\% & 0.17\% & 0.00\% & 0.00\% \\
\midrule

\texttt{ecom\_VN} & 1H & Low    & 0.13\%  & 0.06\% & 0.05\%  & 0.80\% \\
                  &    & Medium & 18.56\% & 1.13\% & 83.48\% & 93.03\% \\

\end{longtable}
}

\subsection{Experiment Details}
\label{app:hyperparams}

All models are implemented in PyTorch~\cite{pytorch} and trained on a single NVIDIA B200 (180GB) GPU. We use the Adam optimizer~\cite{adam}, with the learning rate selected from $\{10^{-3}, 10^{-4}, 5 \times 10^{-5}\}$ via grid search. Across all experiments, the dropout rate is set to $0.2$, training runs for at most 20 epochs with an early stopping patience of 3, $\eta$ is fixed to $1.0$ for $\mathcal{L}_{\text{val}}$, and focal loss is used for $\mathcal{L}_{\text{obs}}$. 

Table~\ref{tab:model_hyperparams} summarizes the dataset-specific hyperparameter settings of \NAME.
We set the input context length ($L$) and prediction horizon ($H$) according to the temporal granularity of each dataset. High-frequency datasets (\texttt{5T}, \texttt{10T}, \texttt{1H}) use a longer window ($L=720, H=336$), whereas lower-frequency datasets (\texttt{1D}, \texttt{1W}, \texttt{1M}) use shorter ranges ($L \in [12, 180], H \in [4, 30]$). The default architecture uses $d_{\text{model}}=128$, $d_{\text{ff}}=256$, and $N_{\text{layers}}=2$. For fully observed settings, we increase the model capacity up to $d_{\text{model}}=256$, $d_{\text{ff}}=1024$, and $N_{\text{layers}}=4$ to better capture dense temporal dependencies.

% Table~\ref{tab:model_hyperparams} summarizes the dataset-specific configurations for \NAME. 
% We align the input context length ($L$) and prediction horizon ($H$) with the temporal 
% resolution: high-frequency datasets (\texttt{5T}, \texttt{10T}, \texttt{1H}) use an extended 
% window ($L=720, H=336$), whereas low-frequency data (\texttt{1D}, \texttt{1W}, \texttt{1M}) 
% utilize shorter ranges ($L \in [12, 180], H \in [4, 30]$). While the default architecture 
% employs $d_{\text{model}}=128$, $d_{\text{ff}}=256$, and $N_{\text{layers}}=2$, we increase 
% the model capacity (up to $d_{\text{model}}=256$, $d_{\text{ff}}=1024$, $N_{\text{layers}}=4$) 
% for fully observed (none-missing) scenarios to better capture dense temporal dependencies.

% \paragraph{Implementation Details.}
% All models are implemented in PyTorch~\cite{pytorch} and trained on a single NVIDIA B200 
% (180GB) GPU. We use the Adam optimizer~\cite{adam}, with the learning rate selected from 
% $\{10^{-3}, 10^{-4}, 5 \times 10^{-5}\}$ via grid search. Across all experiments, the 
% dropout rate is $0.2$, and the training is limited to 20 epochs with an early stopping 
% patience of 3. We set the weight $\eta=1.0$ for $\mathcal{L}_{\text{val}}$ and adopt focal 
% loss for $\mathcal{L}_{\text{obs}}$. 

\begin{table*}[t]
\centering
\small
\setlength{\tabcolsep}{5pt}
\caption{Dataset-specific hyperparameter settings of \NAME for missing-aware forecasting. Datasets and frequencies are presented hierarchically, and rows are merged when multiple missingness settings share the same configuration. Shared hyperparameters are given in the note.}
\label{tab:model_hyperparams}
\resizebox{\textwidth}{!}{%
\begin{tabular}{lll c c c c c c c}
\toprule
\textbf{Dataset} & \textbf{Freq.} & \textbf{Missingness} & \boldmath$d_{\mathrm{model}}$ & \boldmath$n_{\mathrm{heads}}$ & \boldmath$n_{\mathrm{layers}}$ & \boldmath$d_{\mathrm{ff}}$ & \boldmath$L$ & \boldmath$H$ & \textbf{Batch} \\
\midrule

\multirow{3}{*}{Bitbrains-FastStorage}
& 5T & low / medium / high / none & 128 & 8 & 2 & 256  & 720 & 336 & 1024 \\
& \multirow{2}{*}{1H} & low / medium / high & 128 & 8 & 2 & 256  & 336 & 48  & 256  \\
&                     & none                  & 256 & 8 & 4 & 1024 & 336 & 48  & 256  \\
\midrule

\multirow{3}{*}{Bitbrains-Rnd}
& \multirow{2}{*}{5T} & low / high                 & 256 & 16 & 4 & 1024 & 720 & 336 & 1024 \\
&                     & medium                     & 128 & 8  & 2 & 256  & 720 & 336 & 1024 \\
& 1H                  & low / medium / high / none & 128 & 8  & 2 & 256  & 336 & 48  & 256  \\
\midrule

Car Parts
& 1M & high / none & 128 & 8 & 2 & 256 & 12 & 12 & 256 \\
\midrule

\multirow{6}{*}{Electricity}
& \multirow{2}{*}{1D} & low / medium / none  & 128 & 8 & 2 & 256  & 180 & 8   & 256  \\
&                     & high                 & 256 & 8 & 2 & 1024 & 96  & 8   & 256  \\
& \multirow{2}{*}{1H} & low                  & 128 & 8 & 2 & 256  & 720 & 336 & 256  \\
&                     & medium / high        & 128 & 8 & 2 & 256  & 720 & 336 & 1024 \\
& \multirow{2}{*}{1W} & low                  & 128 & 8 & 2 & 256  & 96  & 4   & 64   \\
&                     & medium / high / none & 128 & 8 & 2 & 256  & 96  & 4   & 256  \\
\midrule

Hierarchical Sales
& 1D & low & 256 & 8 & 2 & 1024 & 96 & 96 & 256 \\
\midrule

\multirow{2}{*}{Jena Weather}
& 10T & low & 128 & 8 & 2 & 256 & 720 & 336 & 256 \\
& 1H  & low & 128 & 8 & 2 & 256 & 720 & 336 & 256 \\
\midrule

\multirow{3}{*}{KDD Cup 2018}
& 1H & low / medium / high & 128 & 8 & 2 & 256 & 720 & 336 & 256 \\
& \multirow{2}{*}{1D} & low / medium / none & 128 & 8 & 2 & 256 & 96 & 30 & 256 \\
&                     & high                & 64  & 4 & 1 & 128 & 96 & 30 & 32  \\
\midrule

Restaurant
& 1D & low / medium / high / none & 128 & 8 & 2 & 256 & 48 & 7 & 256 \\
\midrule

\multirow{3}{*}{Temperature Rain}
& \multirow{3}{*}{1D} & low    & 256 & 8 & 4 & 1024 & 336 & 30 & 256 \\
&                     & medium & 64  & 8 & 1 & 256  & 336 & 30 & 256 \\
&                     & none   & 128 & 8 & 2 & 256  & 336 & 30 & 256 \\
\midrule

e\_com\_SB
& 1H & low / high & 128 & 8 & 2 & 256 & 720 & 336 & 256 \\
e\_com\_HN
& 1H & high & 128 & 8 & 2 & 256 & 720 & 336 & 256 \\
e\_com\_AI
& 1H & high & 128 & 8 & 2 & 256 & 720 & 336 & 256 \\
e\_com\_BY
& 1H & medium & 128 & 8 & 2 & 256 & 720 & 336 & 256 \\
e\_com\_BB
& 1H & medium & 128 & 8 & 2 & 256 & 720 & 336 & 256 \\
e\_com\_NR
& 1H & low & 128 & 8 & 2 & 256 & 720 & 336 & 256 \\
e\_com\_NO
& 1H & low / medium & 128 & 8 & 2 & 256 & 720 & 336 & 256 \\
e\_com\_VN
& 1H & low / medium & 128 & 8 & 2 & 256 & 720 & 336 & 256 \\
e\_com\_JP
& 1H & low & 128 & 8 & 2 & 256 & 720 & 336 & 256 \\
e\_com\_HR
& 1H & medium & 128 & 8 & 2 & 256 & 720 & 336 & 256 \\
e\_com\_TA
& 1H & low & 128 & 8 & 2 & 256 & 720 & 336 & 256 \\
e\_com\_ME
& 1H & medium / high & 128 & 8 & 2 & 256 & 720 & 336 & 256 \\
e\_com\_AD
& 1H & low / medium & 128 & 8 & 2 & 256 & 720 & 336 & 256 \\
e\_com\_PL
& 1H & medium & 128 & 8 & 2 & 256 & 720 & 336 & 256 \\
e\_com\_NG
& 1H & medium & 128 & 8 & 2 & 256 & 720 & 336 & 256 \\
e\_com\_BM
& 1H & medium & 128 & 8 & 2 & 256 & 720 & 336 & 256 \\
\bottomrule
\end{tabular}%
}
\vspace{2pt}
\begin{minipage}{\textwidth}
\footnotesize
\textbf{Note.}
Shared hyperparameters across all experiments: dropout $=0.2$, learning rate $=10^{-4}$, epochs $=20$, patience $=3$, $\eta=1$, and $\mathcal{L}_{\text{obs}}$ = focal.
\end{minipage}
\end{table*}

\normalsize
\subsection{Metric Details}
\label{app:metric}

The proposed Observation-Value Joint Entropy (OVJE) metric represents a mathematical formalization of our ``existence precedes value'' philosophy, evaluating observation forecasts together with value estimates at observed positions. To project the regression accuracy into a value-quality scale that can be jointly evaluated with classification, we map the symmetric relative estimation error (sMAPE) $e_t$ to a bounded state quality score $q_t\in(0,1]$:
\begin{equation}
    e_t=\frac{2|y_t-\hat{y}_t|}{|y_t|+|\hat{y}_t|+\epsilon},\quad q_t=\exp(-e_t),
\end{equation}
where $\epsilon=10^{-8}$ and $\hat p_t=\hat p_{\mathrm{obs},t}$ is clipped to $[\epsilon,1-\epsilon]$. Here $o_t=O_t$ and $q_t$ is a quality factor rather than a normalized density. The joint quality score $P_t$ is formulated piecewise:
\begin{equation}
    P_t=\begin{cases}\hat{p}_t\cdot q_t,&o_t=1\\[4pt]1-\hat{p}_t,&o_t=0\end{cases}
\end{equation}

This explicitly unifies the two streams: if the point naturally exists ($o_t=1$), the score accounts for the probability assigned to its existence ($\hat{p}_t$) and the accuracy of its estimated value ($q_t$). Conversely, if the point is missing ($o_t=0$), correctness relies solely on predicting its absence ($1-\hat{p}_t$). This logic compactly reduces to:
\begin{equation}
    P_t=(\hat{p}_t\cdot q_t)^{o_t}\cdot(1-\hat{p}_t)^{1-o_t}.
\end{equation}

Taking the negative logarithm of this joint quality score over all $T$ evaluated positions yields the final OVJE metric:
\begin{equation}
    \text{OVJE}=-\frac{1}{T}\sum_{t=1}^T\log P_t=-\frac{1}{T}\sum_{t=1}^T\left[o_t\cdot\log(\hat{p}_t\cdot q_t)+(1-o_t)\cdot\log(1-\hat{p}_t)\right].
\end{equation}
As shown in \cref{eq:ovje}, this score is exactly the sum of Bernoulli NLL and the observed-position sMAPE contribution, both normalized by the total number of evaluated positions. It cannot replace reporting these components separately, and its value term depends on the observed fraction. Comparisons should therefore use the same evaluation positions and missingness regime. 

Unlike AUC, OVJE is sensitive to probability confidence. For example, take $o=(1,1,0,0)$ and exact value predictions at the two observed positions. Probability vectors $(0.9,0.8,0.2,0.1)$ and $(0.02,0.01,0.009,0.008)$ both yield AUC $=1$ and MAE $=0$, but their OVJE scores are $0.164$ and $2.134$, respectively: the latter assigns very low probability to observations that actually occur. This example illustrates the NLL contribution; the score still decomposes exactly as in \cref{eq:ovje}.

\section{Full Results}

\subsection{Full Missingness Prediction Results}
\label{app:missing_pred_detail}

\cref{tab:full_missing_pred} summarizes the dataset-level results, using the arithmetic mean to aggregate forecasting (\texttt{MAE}, \texttt{MSE}), classification (\texttt{AUC}), and joint (\texttt{OVJE}) metrics. \NAME consistently outperforms baselines, particularly in high-sparsity environments. Notably, on \texttt{ecom\_NG} and \texttt{ecom\_VN}, our model maintains performance despite a massive missingness shift (e.g., from $0.15\%$ during training to $94.43\%$ at test time), whereas competing models prove brittle. These cases suggest sensitivity to observation density in some baselines. High missingness is already visible during validation, however, so they do not establish robustness to an entirely unseen observation process.

\tiny
\setlength{\tabcolsep}{2.2pt}
\begin{longtable}{@{}clcccccccccccc@{}}
\caption{Full missingness prediction Results. Abbreviations: BB-FS = Bitbrains Fast Storage, BB-rnd = Bitbrains Random, Hier. Sales = Hierarchical Sales, Jena Wea. = Jena Weather, KDD Cup = KDD Cup 2018, Temp. Rain = Temperature Rain. Datasets with the same name appearing in multiple row groups differ in sampling frequency. Single-class evaluations use the implementation fallback AUC $=0.500$, for which AUC is mathematically undefined.} \label{tab:full_missing_pred} \\
\toprule
Dataset & Missing Ratio & Metric & \NAME & Crossformer & DLinear & FACT & FEDformer & MICN & OLinear & PatchTST & TimeMixer & WPMixer & iTransformer \\
\midrule
\endfirsthead
\multicolumn{14}{c}{\tablename\ \thetable{} -- Continued from previous page} \\
\toprule
Dataset & Missing Ratio & Metric & \NAME & Crossformer & DLinear & FACT & FEDformer & MICN & OLinear & PatchTST & TimeMixer & WPMixer & iTransformer \\
\midrule
\endhead
\midrule
\multicolumn{14}{r}{Continued on next page} \\
\endfoot
\bottomrule
\endlastfoot
\multirow{16}{*}{\rotatebox{90}{BB-FS}} & \multirow{4}{*}{High} & MAE$\downarrow$ & 0.092 & 0.255 & 0.190 & 0.105 & 0.147 & 0.184 & 0.104 & 0.102 & 0.113 & 0.099 & 0.095 \\*
 &  & MSE$\downarrow$ & 0.702 & 2.904 & 1.202 & 0.773 & 0.738 & 0.860 & 0.717 & 0.747 & 0.721 & 0.727 & 0.709 \\*
 &  & AUC$\uparrow$ & 0.995 & 0.860 & 0.853 & 0.954 & 0.830 & 0.943 & 0.956 & 0.954 & 0.951 & 0.809 & 0.957 \\*
 &  & OVJE$\downarrow$ & 0.378 & 0.803 & 0.919 & 0.487 & 1.179 & 0.835 & 0.462 & 0.485 & 0.495 & 0.710 & 0.505 \\*
\cline{2-14}
\noalign{\penalty10000}
 & \multirow{4}{*}{Medium} & MAE$\downarrow$ & 0.554 & 0.561 & 0.624 & 0.560 & 0.694 & 0.644 & 0.589 & 0.555 & 0.652 & 0.590 & 0.650 \\*
 &  & MSE$\downarrow$ & 0.668 & 0.769 & 0.647 & 0.766 & 0.788 & 0.996 & 0.730 & 0.729 & 0.945 & 0.819 & 0.962 \\*
 &  & AUC$\uparrow$ & 0.995 & 0.820 & 0.783 & 0.932 & 0.384 & 0.826 & 0.906 & 0.939 & 0.973 & 0.440 & 0.931 \\*
 &  & OVJE$\downarrow$ & 0.796 & 1.037 & 1.324 & 0.823 & 1.309 & 0.970 & 0.919 & 1.039 & 0.907 & 1.349 & 0.981 \\
\end{longtable}
\normalsize
\setlength{\tabcolsep}{6pt}

\subsection{Full Value Prediction Results}
Table~\ref{tab:value_prediction_full} presents dataset-level forecasting performance in the absence of the classification head, with \texttt{MAE} and \texttt{MSE} aggregated via the arithmetic mean. \NAME consistently achieves state-of-the-art or competitive results across all regimes. Notably, the catastrophic failure of baseline models on \texttt{ecom\_NG} and \texttt{ecom\_VN} (Medium) persists in this isolated forecasting setting. This recurrence indicates that the value-only architecture contributes to these results, independently of the auxiliary classification objective; it does not imply uniform robustness to arbitrary observation shifts.

\tiny
\setlength{\tabcolsep}{0.6pt}
\begin{longtable}{@{}cclcccccccccccc@{}}
\caption{Value Prediction Results. Abbreviations: BB-FS = Bitbrains Fast Storage, BB-rnd = Bitbrains Random, Hier.\ Sales = Hierarchical Sales, Jena Wea.\ = Jena Weather, Temp.\ Rain = Temperature Rain. Datasets with the same name appearing in multiple row groups differ in sampling frequency.} \label{tab:value_prediction_full} \\
\toprule
Dataset & Frequency & Missing Ratio & Metric & \textbf{Timeflies} & Crossformer & DLinear & FACT & FEDformer & MICN & OLinear & PatchTST & TimeMixer & WPMixer & iTransformer \\
\midrule
\endfirsthead
\multicolumn{15}{c}{\tablename\ \thetable{} -- Continued from previous page} \\
\toprule
Dataset & Frequency & Missing Ratio & Metric & \textbf{Timeflies} & Crossformer & DLinear & FACT & FEDformer & MICN & OLinear & PatchTST & TimeMixer & WPMixer & iTransformer \\
\midrule
\endhead
\midrule
\multicolumn{15}{r}{Continued on next page} \\
\endfoot
\bottomrule
\endlastfoot
\multirow{8}{*}{BB-FS} & \multirow{8}{*}{5T} & \multirow{2}{*}{High} & MAE$\downarrow$ & 0.100 & 0.288 & 0.183 & 0.097 & 0.158 & 0.206 & 0.100 & 0.102 & 0.102 & 0.096 & 0.098 \\*
 &  &  & MSE$\downarrow$ & 0.700 & 3.165 & 1.082 & 0.720 & 0.751 & 1.156 & 0.734 & 0.741 & 0.725 & 0.729 & 0.731 \\*
\cline{4-15}
\noalign{\penalty10000}
 &  & \multirow{2}{*}{Medium} & MAE$\downarrow$ & 0.535 & 0.563 & 0.619 & 0.561 & 0.664 & 0.607 & 0.574 & 0.574 & 0.663 & 0.601 & 0.602 \\*
 &  &  & MSE$\downarrow$ & 0.611 & 0.705 & 0.650 & 0.731 & 0.754 & 0.830 & 0.699 & 0.777 & 0.951 & 0.872 & 0.847 \\*
\cline{4-15}
\noalign{\penalty10000}
 &  & \multirow{2}{*}{Low} & MAE$\downarrow$ & 0.128 & 0.138 & 0.143 & 0.137 & 0.227 & 0.160 & 0.141 & 0.139 & 0.131 & 0.147 & 0.138 \\*
 &  &  & MSE$\downarrow$ & 0.281 & 0.323 & 0.245 & 0.295 & 0.308 & 0.339 & 0.292 & 0.277 & 0.263 & 0.322 & 0.300 \\*
\cline{4-15}
\noalign{\penalty10000}
 &  & \multirow{2}{*}{No} & MAE$\downarrow$ & 0.077 & 0.087 & 0.082 & 0.082 & 0.129 & 0.088 & 0.082 & 0.079 & 0.081 & 0.081 & 0.083 \\*
 &  &  & MSE$\downarrow$ & 0.060 & 0.075 & 0.063 & 0.058 & 0.108 & 0.071 & 0.063 & 0.065 & 0.057 & 0.064 & 0.059 \\
\end{longtable}
\normalsize
\setlength{\tabcolsep}{6pt}

\subsection{Ablation Analysis of Mask-Aware Normalization}
\label{app:mask_aware_norm}
\cref{tab:mask_aware_norm_value} examines the impact of integrating Mask-Aware Normalization (MAN) into observation-aware forecasting baselines. We evaluate three representative models across four missingness regimes. For iTransformer and DLinear, MAN yields negligible performance shifts (e.g., iTransformer stays near 0.325/0.250 under zero missingness). This consistency aligns with our theoretical expectation: when the observation mask is all-ones, MAN reduces to standard Instance Normalization, rendering the forward pass mathematically equivalent. Conversely, OLinear suffers consistent degradation (e.g., MSE increases from 0.301 to 0.367). This is because MAN replaces OLinear's default RevIN module---which utilizes learnable affine parameters---with a purely statistical normalization that lacks sufficient representational capacity. These results underscore that simply swapping normalization layers is insufficient; effective exploitation of missingness patterns requires the dedicated architectural innovations proposed in our method.
\begin{table}[!h]
    \caption{Effect of Mask Aware Normalization on value forecasting performance.
    We compare models without and with mask aware normalization across different missingness regimes.
    Lower MSE and MAE are better. }
    \label{tab:mask_aware_norm_value}
    \centering
    \scriptsize
    \resizebox{\textwidth}{!}{
    \begin{threeparttable}
        \renewcommand{\arraystretch}{1}
        \begin{tabular}{c cc cc cc cc}
            \toprule
            \multirow{2}{*}{Method} &
            \multicolumn{2}{c}{High Missing} &
            \multicolumn{2}{c}{Medium Missing} &
            \multicolumn{2}{c}{Low Missing} &
            \multicolumn{2}{c}{No Missing} \\
            \cmidrule(lr){2-3} \cmidrule(lr){4-5} \cmidrule(lr){6-7} \cmidrule(lr){8-9}
            & w/o MAN & w/ MAN
            & w/o MAN & w/ MAN
            & w/o MAN & w/ MAN
            & w/o MAN & w/ MAN \\
            \midrule
            DLinear
            & 1.716 / 0.369 & 1.724 / 0.387
            & 0.614 / 0.354 & 0.612 / 0.354
            & 1.438 / 0.380 & 1.439 / 0.380
            & 0.367 / 0.283 & 0.367 / 0.282 \\
            
            OLinear
            & 1.586 / 0.311 & 1.959 / 0.406
            & 0.596 / 0.333 & 0.684 / 0.378
            & 1.427 / 0.364 & 1.500 / 0.440
            & 0.301 / 0.240 & 0.367 / 0.311 \\
            
            iTransformer
            & 1.611 / 0.348 & 1.614 / 0.358
            & 0.634 / 0.353 & 0.638 / 0.354
            & 1.462 / 0.379 & 1.465 / 0.380
            & 0.326 / 0.251 & 0.325 / 0.250 \\
            \bottomrule
        \end{tabular}
    \end{threeparttable}
    }
\end{table}

\cref{tab:mask_aware_norm_missing} presents an ablation study of Mask-Aware Normalization (MAN) within our missing-aware framework. We observe several key insights. First, for iTransformer under the zero-missingness regime, MAN and standard instance normalization yield nearly identical results (MSE: 0.324 vs. 0.326), validating our theoretical assertion that MAN reduces to standard normalization when the observation mask is all-ones. Second, under non-trivial missingness, MAN consistently enhances MAE and OVJE for iTransformer (e.g., MAE decreases from 0.354 to 0.338 under high missingness). This suggests that restricting statistics to observed entries provides a more robust, unbiased normalization signal. Third, OLinear reveals a trade-off between learnability and missingness-awareness: while MAN improves MAE under high missingness (0.328 vs. 0.315), it degrades MSE in the no-missing case (0.317 vs. 0.377) by replacing RevIN's learnable affine parameters with purely statistical normalization. Finally, DLinear shows mixed, metric-dependent effects: under high missingness, MSE improves from 1.714 to 1.614 while OVJE worsens from 1.069 to 1.124; under medium missingness, MAE improves from 0.367 to 0.346 while MSE worsens from 0.611 to 0.634. Under no missingness, changes are small. Thus, MAN does not improve every metric or regime.
\begin{table*}[!tb]
    \caption{Effect of Mask-Aware Normalization (MAN) on missing-aware forecasting. We compare three baselines with and without MAN across four missingness regimes. Metrics: MSE$\downarrow$ and MAE$\downarrow$ evaluate value prediction accuracy; AUC$\uparrow$ measures observation classification quality (observed=1); OVJE$\downarrow$ (Obs-Value Joint sMAPE Log Loss) captures joint forecasting performance. "--" indicates AUC is undefined when no missing values exist. Results are averaged across all datasets and prediction horizons.}
    \label{tab:mask_aware_norm_missing}
    \centering
    \scriptsize
    \resizebox{\textwidth}{!}{
    \begin{threeparttable}
        \renewcommand{\arraystretch}{1.15}
        \begin{tabular}{c c cccc cccc cccc cccc}
            \toprule
            \multirow{2}{*}{Method} & \multirow{2}{*}{Setting}
            & \multicolumn{4}{c}{High Missing}
            & \multicolumn{4}{c}{Medium Missing}
            & \multicolumn{4}{c}{Low Missing}
            & \multicolumn{4}{c}{No Missing} \\
            \cmidrule(lr){3-6} \cmidrule(lr){7-10} \cmidrule(lr){11-14} \cmidrule(lr){15-18}
            &  & MSE$\downarrow$ & MAE$\downarrow$ & AUC$\uparrow$ & OVJE$\downarrow$
               & MSE$\downarrow$ & MAE$\downarrow$ & AUC$\uparrow$ & OVJE$\downarrow$
               & MSE$\downarrow$ & MAE$\downarrow$ & AUC$\uparrow$ & OVJE$\downarrow$
               & MSE$\downarrow$ & MAE$\downarrow$ & AUC$\uparrow$ & OVJE$\downarrow$ \\
            \midrule
            \multirow{2}{*}{DLinear}
                & w/o MAN & 1.714 & 0.383 & 0.638 & 1.069 & 0.611 & 0.367 & 0.528 & 1.328 & 1.433 & 0.387 & 0.507 & 1.245 & 0.371 & 0.286 & -- & 1.020 \\
                & w/ MAN  & 1.614 & 0.324 & 0.527 & 1.124 & 0.634 & 0.346 & 0.539 & 1.271 & 1.443 & 0.368 & 0.505 & 1.082 & 0.372 & 0.289 & - & 1.014 \\
            \midrule
            \multirow{2}{*}{OLinear}
                & w/o MAN & 1.622 & 0.328 & 0.740 & 0.762 & 0.608 & 0.352 & 0.623 & 1.475 & 1.438 & 0.376 & 0.525 & 0.958 & 0.317 & 0.247 & -- & 0.695 \\
                & w/ MAN  & 1.609 & 0.315 & 0.760 & 0.751 & 0.633 & 0.350 & 0.614 & 1.329 & 1.445 & 0.366 & 0.537 & 0.827 & 0.377 & 0.267 & - & 0.664 \\
            \midrule
            \multirow{2}{*}{iTransformer}
                & w/o MAN & 1.590 & 0.354 & 0.680 & 0.846 & 0.638 & 0.367 & 0.566 & 1.263 & 1.452 & 0.387 & 0.524 & 1.017 & 0.324 & 0.251 & -- & 0.730 \\
                & w/ MAN  & 1.655 & 0.338 & 0.742 & 0.816 & 0.652 & 0.356 & 0.591 & 1.088 & 1.458 & 0.379 & 0.494 & 0.880 & 0.326 & 0.251 & - & 0.737 \\
            \bottomrule
        \end{tabular}
    \end{threeparttable}
    }
\end{table*}

% ------------------------------------------------------------

\subsection{Additional interpretation of observation-guided attention}
\label{app:attention_case}
To analyze how \NAME handles irregular missingness, we visualize its internal attention dynamics on a real-world sample. As shown in \cref{fig:model_visualization}(a), the model accurately recovers and extrapolates the underlying trend despite severe discontinuity in the historical context, demonstrating its ability to extract coherent temporal structure from highly fragmented inputs.
\cref{fig:model_visualization}(b)--(e) further reveal how the conditioned attention is formed. The value attention in \cref{fig:model_visualization}(b) shows broad activation over the timeline, capturing general temporal dependencies but remaining vulnerable to noisy missing patches. In contrast, the observation-aware bias in \cref{fig:model_visualization}(c) and the pairwise reliability attention in \cref{fig:model_visualization}(d) encode structured patterns aligned with the missingness distribution. Their fusion in \cref{fig:model_visualization}(e) yields selective routing, suppressing unreliable sparse keys while emphasizing informative patches. This reliability-aware filtering limits the propagation of missing-induced noise and explains the stable trend recovery in \cref{fig:model_visualization}(a).
\cref{fig:model_visualization}(f) visualizes the complementary missingness probabilities $\hat{p}_{\mathrm{miss}}=1-\hat{p}_{\mathrm{obs}}$. Higher plotted probabilities correspond to future missing events; $\hat{p}_{\mathrm{obs}}=\sigma(\hat{O})$ instead denotes observation probability. This distinction is used consistently in the loss and OVJE, with observed positions as the positive class.

\section{Controlled Missingness and Contribution Analysis}
\label{app:controlled}

\subsection{Random, structured, and shifted missingness}
\label{app:controlled_robustness}
We use five complete datasets: Electricity H, Bitbrains FS, Restaurant, US Births, and Saugeen Day. Identical masks are shared across variants within each dataset, with the same splits and training protocol; underlying complete evaluation targets are retained for value-error evaluation. IID-10 and IID-50 independently mask 10\% and 50\% of positions in all splits. BLOCK-50 uses periodic missing blocks at a 50\% rate, constructed independently of the underlying values. SHIFT-10-50 uses 10\% IID missingness in training/validation and 50\% at test time. Each result is an equal-weight arithmetic average of dataset-level metrics; these controlled results use a different evaluation protocol from naturally missing targets in \BenchName.

Variant A is value-only PatchTST; B adds mask-aware normalization; D incorporates historical-observation modeling and residual fusion; E additionally uses future-observation supervision. Table~\ref{tab:controlled_robustness} reports their MAE/MSE and the full model's observation AUC/Brier.

\begin{table}[!ht]
\centering
\caption{Controlled missingness on five complete datasets. Value cells report MAE/MSE; the final column reports E's observation ROC-AUC/Brier. Lower MAE, MSE, and Brier are better; higher AUC is better.}
\label{tab:controlled_robustness}
\scriptsize
\setlength{\tabcolsep}{3pt}
\resizebox{\textwidth}{!}{%
\begin{tabular}{lccccc c}
\toprule
Setting & Train/Val/Test & A: PatchTST & B: Mask-aware & D: Hist-Obs & E: Full & E AUC/Brier \\
\midrule
IID-10 & 10\%/10\%/10\% & 0.542/1.165 & 0.543/1.169 & 0.524/1.144 & 0.521/1.142 & 0.494/0.113 \\
IID-50 & 50\%/50\%/50\% & 0.570/1.156 & 0.562/1.146 & 0.539/1.111 & 0.538/1.110 & 0.507/0.266 \\
BLOCK-50 & 50\% periodic & 0.812/2.517 & 0.705/2.316 & 0.706/2.317 & 0.704/2.323 & 0.930/0.079 \\
SHIFT-10-50 & 10\%/10\%/50\% & 0.632/1.282 & 0.612/1.252 & 0.650/1.327 & 0.652/1.331 & 0.504/0.335 \\
\bottomrule
\end{tabular}}
\end{table}

Under IID masks, E improves average value accuracy over A while observation AUC remains near chance. The historical mask identifies valid inputs rather than encoding predictable observation dynamics. BLOCK-50 makes observation prediction learnable (AUC $0.930$), but E's MSE is slightly worse than B's ($2.323$ versus $2.316$). Under SHIFT-10-50, B performs best and E is worse than both A and B in MAE/MSE. Thus, the in-distribution results do not imply robustness to a previously unseen observation process.

\subsection{Separating architecture, historical inputs, and future supervision}
\label{app:controlled_decomposition}
Using the same five datasets and controlled protocol, we add No-Missing as a negative control. All variants share masks, initialization of shared components, and training budgets. We compare:
\begin{itemize}[leftmargin=*,nosep]
\item \textbf{M:} a parameter-matched value-only backbone with no observation stream or future-observation supervision.
\item \textbf{V0 (Neutral-O):} the generic Timeflies architecture with a constant neutral observation representation. The true historical mask is retained only for observed-entry normalization; future-observation supervision is disabled.
\item \textbf{V1 (Hist-O only):} V0 with true historical masks, intervals, and patch reliability, still without future-observation supervision.
\item \textbf{V2 (Shuffled Aux-O):} V1 with the full auxiliary head, trained on shuffled future-observation labels preserving the marginal observation rate.
\item \textbf{V3 (Full):} the same architecture and auxiliary head with correctly aligned future-observation labels.
\end{itemize}
The comparisons distinguish generic architecture (M to V0), historical input-validity information (V0 to V1), non-informative auxiliary supervision (V1 to V2), and label alignment (V2 to V3).

\begin{table}[!ht]
\centering
\caption{Controlled contribution analysis. M and V0--V3 cells report MAE/MSE, averaged arithmetically with equal weight across the five datasets. AUC uses observed=1; it is undefined (N/A) under No-Missing.}
\label{tab:controlled_decomposition}
\scriptsize
\setlength{\tabcolsep}{3pt}
\resizebox{\textwidth}{!}{%
\begin{tabular}{lccccc cc}
\toprule
Regime & M & V0 & V1 & V2 & V3 & V2 AUC/Brier & V3 AUC/Brier \\
\midrule
No-Missing & 0.535/1.140 & 0.520/1.115 & 0.520/1.115 & 0.519/1.113 & 0.519/1.113 & N/A/0.000 & N/A/0.000 \\
IID-10 & 0.543/1.169 & 0.532/1.156 & 0.524/1.144 & 0.523/1.143 & 0.521/1.142 & 0.501/0.114 & 0.494/0.113 \\
IID-50 & 0.562/1.146 & 0.550/1.129 & 0.539/1.111 & 0.539/1.111 & 0.538/1.110 & 0.499/0.267 & 0.507/0.266 \\
BLOCK-50 & 0.705/2.316 & 0.709/2.324 & 0.706/2.317 & 0.711/2.340 & 0.704/2.323 & 0.501/0.249 & 0.930/0.079 \\
\bottomrule
\end{tabular}}
\end{table}

Under No-Missing, V0=V1 and V2=V3: constant masks contain no nontrivial observability information. The M-to-V0 improvement therefore reflects generic architecture, with a small auxiliary optimization effect thereafter. Under IID-10/50, V0-to-V1 gains reflect input-validity awareness under MCAR, not state-related observation dynamics. Under BLOCK-50, correct instead of shuffled labels raise AUC from $0.501$ to $0.930$ and lower Brier from $0.249$ to $0.079$, establishing that the model learns the structured observation process. Yet V3 and M have nearly identical value accuracy ($0.704/2.323$ versus $0.705/2.316$ MAE/MSE). Predictable observability alone is therefore insufficient to improve state estimation when missingness is value-independent. Joint modeling supplies an additional observation forecast; its incremental value-forecasting benefit is data- and mechanism-dependent, rather than guaranteed under MCAR or value-independent masks.

\section{Limitations}
\label{app:limitations}
\begin{enumerate}[leftmargin=*,nosep]
    \item \textbf{Constant overhead:} The observation-aware modules incur computational overhead even when all data is observed. An adaptive gating mechanism that bypasses these modules when missingness is negligible could improve efficiency.
    \item \textbf{Channel independence assumption:} Following PatchTST, \NAME processes each channel independently. Cross-channel missing patterns (e.g., correlated sensor failures) are not explicitly modeled, though the observation mask implicitly captures some of this information.
    \item \textbf{Observation-process transferability:} Historical observation information must remain useful at test time. IID masks can provide input-validity cues without predictable observation dynamics; predictable but value-independent masks need not improve value accuracy. Previously unseen shifts in the marginal missing rate or conditional observation mechanism can degrade performance (\appref{app:controlled}), so the natural-shift results are not general robustness guarantees.
    \item \textbf{Evaluation scope:} We focus on \BenchName with natural missingness, supplemented by controlled masks on five complete datasets. Evaluation on additional domains (e.g., clinical time series, financial time series \cite{hu2025finmamba}) would strengthen generalizability claims.
\end{enumerate}

\section{Broader Impact}
\label{app:broader_impact}
Missing-aware time series forecasting has broad applications in domains where sensor failures, communication dropouts, or irregular sampling create incomplete observations---including energy grid management, environmental monitoring, healthcare, and cloud infrastructure management. By providing both value forecasts and observation-probability estimates, \NAME enables downstream decision-making systems to incorporate observation uncertainty. We do not foresee negative societal impacts specific to this work beyond those inherent to time series forecasting in general.

% \newpage
% \input{checklist.tex}

\end{document}